\documentclass[10pt,twocolumn,letterpaper]{article}

\usepackage{iccv}
\usepackage{times}
\usepackage{epsfig}
\usepackage{graphicx}
\usepackage{amsmath}
\usepackage{amssymb}
\usepackage{subcaption}
\usepackage{pifont}
\usepackage{tabularx}
\usepackage{arydshln}
\usepackage{booktabs}
\usepackage{multirow}
\usepackage{tabu}
\usepackage{siunitx}
\usepackage[symbol]{footmisc}

\usepackage[dvipsnames,svgnames,x11names]{xcolor}

\usepackage[pagebackref=true,breaklinks=true,letterpaper=true,colorlinks,bookmarks=false]{hyperref}

\iccvfinalcopy

\ificcvfinal\fi

\begin{document}

\title{\vspace{-12mm} Self-Supervised Object Detection via Generative Image Synthesis \vspace{-6mm}}

\author{
Siva Karthik Mustikovela\textsuperscript{1,3}\thanks{Siva Karthik Mustikovela was an intern at NVIDIA during the project.} \hspace{3mm}
Shalini De Mello\textsuperscript{1} \hspace{2mm} Aayush Prakash\textsuperscript{1} \\
Umar Iqbal\textsuperscript{1} \hspace{2mm}
Sifei Liu\textsuperscript{1} \hspace{2mm}
Thu Nguyen-Phuoc\textsuperscript{2} \hspace{2mm}
Carsten Rother\textsuperscript{3} \hspace{2mm}
Jan Kautz\textsuperscript{1} \\
\vspace{1mm}
\textsuperscript{1}NVIDIA \qquad \textsuperscript{2}University of Bath \qquad \textsuperscript{3}Heidelberg University\\ 
{\tt\small \{siva.mustikovela, carsten.rother\}@iwr.uni-heidelberg.de; aayush382.iitkgp@gmail.com;} \\ 
{\tt\small T.Nguyen.Phuoc@bath.ac.uk; \{shalinig, sifeil, uiqbal, jkautz\}@nvidia.com}
}

\maketitle
\ificcvfinal\thispagestyle{empty}\fi

\vspace{-4mm}
\begin{abstract}
\vspace{-2mm}
We present SSOD -- the first end-to-end analysis-by-synthesis framework with controllable GANs for the task of self-supervised object detection. We use collections of real-world images without bounding box annotations to learn to synthesize and detect objects. We leverage controllable GANs to synthesize images with pre-defined object properties and use them to train object detectors. We propose a tight end-to-end coupling of the synthesis and detection networks to optimally train our system. Finally, we also propose a method to optimally adapt SSOD to an intended target data without requiring labels for it. For the task of car detection, on the challenging KITTI and Cityscapes datasets, we show that SSOD outperforms the prior state-of-the-art purely image-based self-supervised object detection method Wetectron. Even without requiring any 3D CAD assets, it also surpasses the state-of-the-art rendering-based method Meta-Sim2. Our work advances the field of self-supervised object detection by introducing a successful new paradigm of using controllable GAN-based image synthesis for it and by significantly improving the baseline accuracy of the task. We open-source our code at \url{https://github.com/NVlabs/SSOD}.
\end{abstract}
\vspace{-5mm}
\section{Introduction}
\vspace{-1mm}
Object detection plays a crucial role in various 
autonomous vision pipelines, \textit{e.g.}, in robotics and self-driving.
Convolutional neural networks-based detection methods, such as~\cite{renNIPS15fasterrcnn, Lin_2017_FPN}, have achieved impressive performance. However, they are fully-supervised and require large amounts of human annotated data, which is time-consuming to acquire for all object types and operating environments. They also do not scale well when target domains change, \textit{e.g.}, from one city to another in self-driving.

To reduce annotations, some existing works train detectors without requiring bounding box annotations 
and follow two paradigms. The first is of self/weakly supervised object detection methods~\cite{ren-cvpr020_wetectron1, ren-eccv2020_wetectron2, Zeng_2019_ICCV_wsod2}, which either use image-level object presence labels (a.k.a. self-supervision) or point/scribble annotations (a.k.a weak-supervision). They also rely on high-quality object proposals detected by methods requiring human annotations~\cite{zitnick2014edgeboxes}.
The second paradigm is of rendering-based methods, including Meta-Sim~\cite{kar2019metasim} and Meta-Sim2~\cite{devaranjan2020metasim2}, which learn object detection from synthetically rendered images. Creating them, however, requires large collections of high-quality 3D CAD models for all the objects in the scene, manual scene setups and expensive rendering engines. Such images also tends to have a large domain gap from real-world ones.

\begin{figure}
	\centering
	\includegraphics[width=\linewidth]{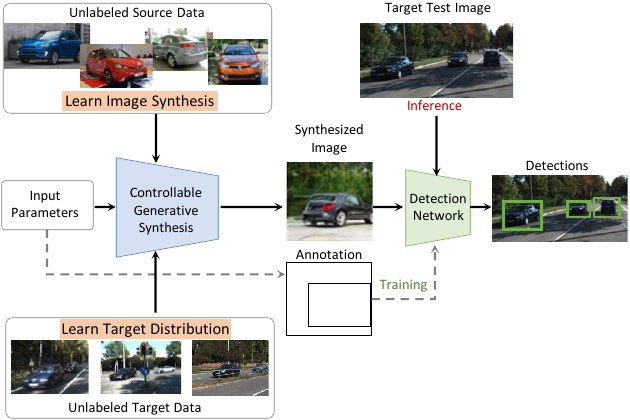}
	\caption{\textbf{Self-Supervised Object Detection.} We learn object detection purely using natural image collections without bounding box labels. We leverage controllable GANs to synthesize images and to detect objects together in a tightly coupled framework. We learn image synthesis from unlabeled singe-object source images (\textit{e.g.}, Compcars~\cite{yang2015large}) and optimally adapt our framework to any multi-object unlabeled target dataset (\textit{e.g.}, KITTI~\cite{Geiger2012CVPR_kitti}).}
	\label{fig:teaser}
	\vspace{-5mm}
\end{figure}

Recently, there has been much progress in making Generative Adversarial Networks (GANs)~\cite{NIPS2014_gans} controllable using input parameters like shape, viewpoint, position and keypoints~\cite{nguyenhologan, BlockGAN2020, niemeyer2020giraffe, schwarz2020graf},
opening up the possibility of synthesizing images with desired attributes. Controllable GANs have also been used successfully to learn other vision tasks, \textit{e.g.}, viewpoint~\cite{mustikovelaCVPR20_ssv} and keypoints~\cite{thewlisICCVT2017, zhang2018unsupervised, tomas2018neurips} estimation in a self-supervised manner, but have not been explored previously for self-supervised object detection. 

Inspired by these, we propose the first end-to-end analysis-by-synthesis framework for self-supervised object detection using controllable GANs, called SSOD (Fig.~\ref{fig:teaser}). We learn to both synthesize images and detect objects purely using unlabeled image collections, \textit{i.e.}, without requiring bounding box-labels and without using 3D CAD assets -- a multi-faceted challenge not addressed previously. We learn a generator for object image synthesis using real-world single-object image collections without bounding box labels. By leveraging controllable GANs, which provide control over the 3D location and orientation of an object, we also obtain its corresponding bounding box annotation. To optimally train SSOD, we tightly couple the synthesis and detection networks in an end-to-end fashion and train them jointly. Finally, we learn to optimally adapt SSOD to a multi-object target dataset, also without requiring labels for it and improve accuracy further.
 
 We validate SSOD on the challenging KITTI~\cite{Geiger2012CVPR_kitti} and Cityscapes~\cite{Cordts_2016_CVPR_cityscapes} datasets for car object detection. SSOD outperforms the best prior image-based self-supervised object detection method Wetectron~\cite{ren-eccv2020_wetectron2} with significantly better detection accuracy. Furthermore, even without using any 3D CAD assets or scene layout priors it also surpasses the best rendering-based method Meta-Sim2~\cite{devaranjan2020metasim2}. 
To the best of our knowledge, SSOD is the first work to explore using controllable GANs for fully self-supervised object detection. Hence, it opens up a new paradigm for advancing further research in this area. SSOD significantly outperforms all competing image-based methods and serves as a strong baseline for future work.

To summarize, our main contributions are:
\begin{itemize}
    \vspace{-2mm}
    \item We propose a novel self-supervised object detection framework via controllable generative synthesis, which uses only image collections without any kind of bounding box annotations. 

    \vspace{-1mm}
    \item We propose an end-to-end analysis-by-synthesis framework, which can optimally adapt the synthesizer to both the downstream task of object detection and to a target dataset in a purely self-supervised manner. 
    
    \vspace{-1mm}
    \item Our experiments on two real-world datasets show $\sim$2x performance improvement over SOTA image-based self-supervised object detection methods.
    Also, without using 3D CAD assets, SSOD outperforms on average, the rendering-based baseline of Meta-Sim2~\cite{devaranjan2020metasim2}. 
    
\end{itemize}
\vspace{-3mm}
\section{Related Work}
\vspace{-1mm}
\paragraph{Self-supervised task learning.} 
Several recent works attempt to learn a variety of 2D and 3D computer vision tasks in a self-supervised manner. In 2D computer vision, several works tackle the problem of object keypoint estimation~\cite{thewlisICCVT2017, zhang2018unsupervised, tomas2018neurips} and part segmentation~\cite{hung2019scops, collins2018deep}. \cite{seggan_bielski} obtains an object mask along with the generated image. However, there is no control over the pose and style of the generated object. Alongside, in 3D computer vision, there are several attempts to learn object reconstruction~\cite{cmrKanazawa18, kulkarni2019csm, umr2020,vmr2020}, viewpoint estimation~\cite{mustikovelaCVPR20_ssv} and point cloud estimation~\cite{navaneet2020ssl3drecon}. These works present interesting approaches to address their respective problems for single object images, but do not address multi-object analysis. 

Concurrently, there has also been tremendous progress in high-quality controllable generative synthesis using learned 3D object representations~\cite{BlockGAN2020, nguyenhologan, niemeyer2020giraffe, schwarz2020graf, EhrhardtGrothEtAl:RELATE:NeurIPS:2020} or implicit representations~\cite{mildenhall2020nerf, kaizhang2020, Riegler2020SVS}. Some of these works have been used in analysis-by-synthesis frameworks to solve computer vision tasks, including 3D reconstruction~\cite{umr2020, vmr2020, henderson20cvpr, henderson19ijcv}, viewpoint estimation~\cite{mustikovelaCVPR20_ssv} and keypoint estimation~\cite{tomas2018neurips}.
However, no prior work explores self-supervised object detection via controllable GANs and we are the first work to do so.

\vspace{-5mm}
\paragraph{Weakly supervised object detection.} 
Recent works also address the problem of self-supervised object detection using only a collection of images and image-level tags of object presence. Such methods pose the problem either in a multiple instance~\cite{bilen_wsod, tang2018pcl, Zeng_2019_ICCV_wsod2, ren-cvpr020_wetectron1, Gao_2019_ICCV}, discriminative~\cite{Shen_2018_CVPR}, curriculum~\cite{zigzag_wsod, kantorov2016, ren-eccv2020_wetectron2} or self-taught~\cite{Jie_2017_CVPR} learning framework. However, such methods rely heavily on object proposals generated by methods like ~\cite{zitnick2014edgeboxes, APBMM2014_mcg, Uijlings13_selsearch}, which, themselves, need low-level edge-based annotations from humans. Additionally, they also cannot modify or control input images according to the requirements of the detector or a target dataset. In contrast we learn a controllable synthesis module, to synthesize images that maximize the detector's performance on a target dataset.

\vspace{-5mm}
\paragraph{Learning object detection from synthetic data.} 
Works like~\cite{Richter_2017, cabon2020vkitti2, gaidon2016virtual, kar2019metasim, devaranjan2020metasim2, Prakash2019StructuredDR} learn object detection through synthetic data from graphics renderers. \cite{Richter_2017} obtains synthetic images and annotations from a game engine. In~\cite{cabon2020vkitti2, gaidon2016virtual}, the authors exactly mimic real world dataset (\textit{e.g.}, KITTI~\cite{Geiger2012CVPR_kitti}) in a synthetic simulator. In~\cite{Prakash2019StructuredDR}, the authors synthesize scenes by randomizing location, orientations and textures of objects of interest in a scene. In Meta-Sim~\cite{kar2019metasim} and Meta-Sim2~\cite{devaranjan2020metasim2}, the authors propose a strategy to learn optimal scene parameters to generate images similar to a target dataset. 
While methods like~\cite{cabon2020vkitti2, gaidon2016virtual} use annotations from real world datasets to mimic the datasets in synthetic worlds, other methods like~\cite{Richter_2017, devaranjan2020metasim2, Prakash2019StructuredDR} generate synthetic data without using any real world annotations. 
While these approaches learn only from rendered data, they require 3D CAD models of objects and scenes along with rendering setups, both of which are expensive to acquire. Moreover, graphics renderers are often not differentiable making it difficult to learn and propagate gradients through them for learning a downstream task. Also, synthetic data introduces a domain gap with respect to real target data both in terms of appearance and layout of scenes that affects detection accuracy. 
In contrast, our goal is to learn both data generation and object detection from real-world images without bounding box annotations and without requiring 3D CAD models or rendering setups. Our GAN-based framework allows us to adapt to the distribution of the target data and synthesize data that is optimal for the downstream task.
\begin{figure*}[h]
	\centering
	\includegraphics[width=\textwidth]{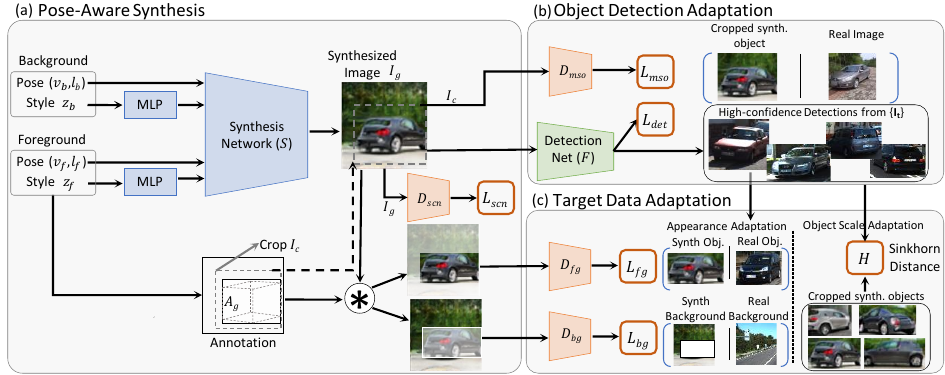}
	\caption{\textbf{Overview of Self-Supervised Object Detection.}
SSOD contains three modules: (a) a pose-aware synthesis module that generates images with objects in pre-defined poses using a controllable GAN for training object detectors; (b) an object detection adaptation module that guides the synthesis process to be optimal for the downstream task of object detection and the (c) a target data adaption module that helps SSOD to adapt optimally to a target data distribution. We train all modules in a tightly-coupled end-to-end manner.		}
\label{fig:approach_overview}
\vspace{-4mm}
\end{figure*}

\vspace{-2mm}
\section{Self-Supervised Object Detection}
\vspace{-2mm}

\subsection{Problem Setup}
\vspace{-1mm}

Our goal is to learn a detection network $\mathcal{F}$, which best detects objects (\textit{e.g.}, cars) in a target domain (\textit{e.g.}, outdoor driving scenes from a city). We further assume that we have available to us an unlabeled image collection $\{\textbf{I}_{t}\}$ from the target domain each containing an unknown number of objects per image (see examples in Fig.~\ref{fig:teaser}). To train $\mathcal{F}$, we leverage object images and their bounding box annotations synthesized by a controllable generative  network $\mathcal{S}$, which, in turn, is also learnt using unlabeled object collections. Specifically, to learn $\mathcal{S}$, we use an additional sufficiently large unlabeled (bounding box annotation free)  single-object source collection $\{\textbf{I}_{s}\}$, containing images with only one object per image, but not necessarily from the target domain where the detector must operate (see examples in  Fig.~\ref{fig:teaser}). We discuss more about the need for this assumption in Sec~\ref{sec:pas}. We train our system with both $\{\textbf{I}_{t}\}$ and $\{\textbf{I}_{s}\}$, and evaluate it on a held-out labeled \textit{validation} set from the target domain, which is disjoint from $\{\textbf{I}_{t}\}$ and is never used for training.

\subsection{Overview of SSOD}\label{sec:overview}
\vspace{-1mm}
We present an overview of SSOD in Fig.~\ref{fig:approach_overview}. It contains three modules: (a) a pose-aware synthesis; (b) an object detection adaptation and (c) a target data adaption module.

The pose-aware synthesis module (Fig.~\ref{fig:approach_overview}(a)) contains a controllable synthesis network $\mathcal{S}$. We model $\mathcal{S}$ by a pose-aware generator, which synthesizes images $\{\textbf{I}_{g}\}$ of objects conditioned on the pose parameters (viewpoint ($\boldsymbol{v}$) and location ($\boldsymbol{l}$)) and obtain 2D bounding box annotations $\{\textbf{A}_{g}\}$ for them. Using the synthesized image-annotation pairs $\langle\textbf{I}_{g}, \textbf{A}_{g}\rangle$, along with images from $\{\textbf{I}_{t}\}$, we train the object detector $\mathcal{F}$. 
The object detection adaptation module (Fig.~\ref{fig:approach_overview}(b)) is designed to provide feedback to the synthesis network $\mathcal{S}$ to optimally adapt it to the downstream task of object detection. It tightly couples the object detector $\mathcal{F}$ and synthesizer $\mathcal{S}$ for joint end-to-end training and also introduces specific losses to guide the synthesis process towards better object detection learning. 

Lastly, the target data adaptation module (Fig.~\ref{fig:approach_overview}(c)) helps reduce the domain gap between the images synthesized by $\mathcal{S}$ and those in the target domain $\{\textbf{I}_{t}\}$. It does so by introducing a set of spatially localized discriminative networks, which adapt the synthesis network $\mathcal{S}$ towards generating images closer to the target data distribution in terms of overall image appearance and scale of objects.

 \noindent We train SSOD in two stages -- uncoupled and coupled. During uncoupled training, we pre-train the synthesis network $\mathcal{S}$ on $\{\textbf{I}_{s}\}$ without feedback from other modules. Next, we synthesize image-annotation pairs with $\mathcal{S}$ and use them along with $\{\textbf{I}_{t}\}$ to pre-train $\mathcal{F}$. 
During the next coupled training phase, we jointly fine-tune SSOD's modules with both the source $\{\textbf{I}_{s}\}$ and target $\{\textbf{I}_{t}\}$ images, and the data synthesized by $\mathcal{S}$. We alternatively train $\mathcal{S}$ in one iteration and all other networks in the next one. We describe all the modules of SSOD in detail in the following sections.

\vspace{-1mm}
\subsection{Pose-Aware Synthesis}\label{sec:pas}
\label{sec:synth_net}
\begin{figure}[ht!]
	\centering
	\includegraphics[width=\columnwidth]{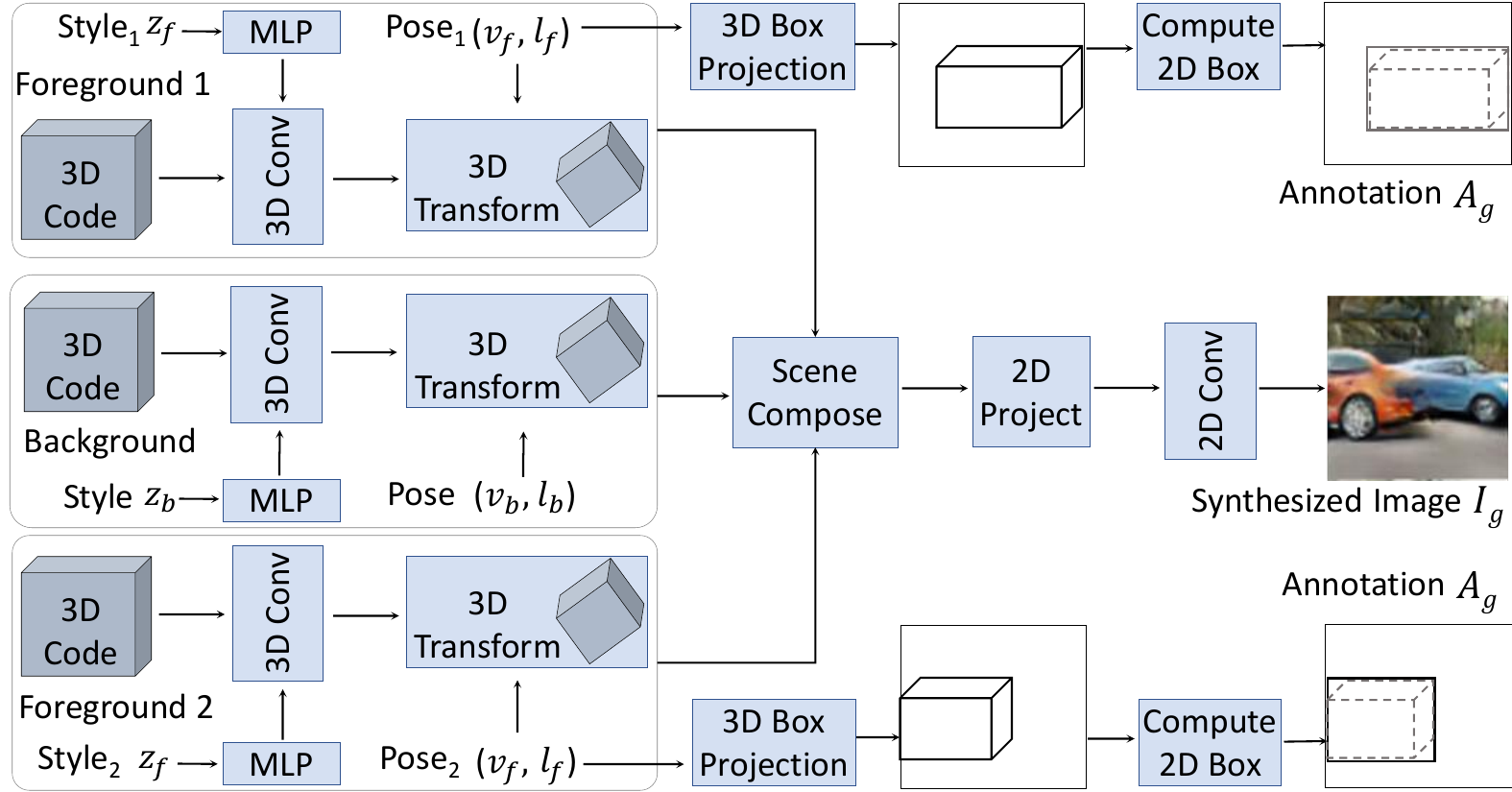}
	\caption{\textbf{Pose-Aware Synthesis Network ($\mathcal{S}$) Overview.} $\mathcal{S}$ takes as input separate style codes ($\boldsymbol{z}$) and poses ($\boldsymbol{v}$, $\boldsymbol{l}$) for the background and one/more foreground objects; transforms their respective learned 3D codes with the provided poses; and synthesizes images after passing them through several 3D convolutional, 2D projection and 2D convolutional layers. We use the provided poses to compute 2D bounding box labels for the synthesized objects.}
	\label{fig:synthesis_net}
	\vspace{-6mm}
\end{figure}

\vspace{-2mm}
Our pose-aware synthesis network $\mathcal{S}$ is inspired by the recent BlockGAN~\cite{BlockGAN2020}, which has several desirable properties for object detection. It allows control over style, pose and number of objects in the scene by disentangling the background and foreground objects. Its architecture is illustrated in Fig.~\ref{fig:synthesis_net}. To make BlockGAN~\cite{BlockGAN2020} amenable to target data adaptation, we augment it with MLP blocks which learn to modify style vectors for both the foreground and background before they are input to the generator, such that the synthesized images are closer to the target dataset (Fig.~\ref{fig:synthesis_net}). 

The synthesis network $\mathcal{S}$  generates a scene $I_g$ containing the foreground object in the specified location and orientation. The network contains category specific learnable canonical 3D codes for foreground and background objects, which are randomly initialized and updated during training. The 3D latent code of each object is passed through a corresponding set of 3D convolutions where the style of the object is controlled by input 1D style code vectors (from a uniform distribution) $\boldsymbol{z}_f$ for the foreground and $\boldsymbol{z}_b$ for the background through AdaIN (Fig.~\ref{fig:synthesis_net}). These 3D features are further transformed using their input poses ($\boldsymbol{v}_f$, $\boldsymbol{l}_f$) for one or more foreground objects. The value of $\boldsymbol{v}_f$ represents azimuth of the object and  $\boldsymbol{l}_f$ represents its horizontal and depth translation. Each object is processed separately in its own 3D convolution branch. The resulting 3D features of all objects are collated using an element-wise maximum operation and then projected onto 2D using a perspective camera transformation followed by a set of 2D convolutions to yield $I_g$. The original BlockGAN~\cite{BlockGAN2020} generates images at a resolution of $64 \times 64$. For our $\mathcal{S}$, we modify it and adopt the strategy of progressive growing of GANs~\cite{style_gan, karras2018progressive} to increase its synthesis resolution to $256 \times 256$.

We train $\mathcal{S}$ in a GAN setup with an adversarial loss~\cite{arjovsky2017wasserstein}  $\mathcal{L}_\mathit{scn}$ computed using a scene discriminator $\mathcal{D}_{scn}$ as: 

\begin{equation}
	\mathcal{L}_\mathit{scn} = -\mathbb{E}_{I_g \sim p_\mathrm{synth}}[\mathcal{D}_{scn}(I_g)],
	\label{eqn:scn_loss}
\end{equation}
\noindent where $\mathcal{D}_{scn}(I_g)$ is the class membership score predicted by the scene discriminator $\mathcal{D}_{scn}$ for a synthesized image. This is one among several losses that we use to train $\mathcal{S}$. The real images input to $\mathcal{D}_{scn}$ are sampled from $\{\textbf{I}_s\}$.

To train $\mathcal{S}$, we use a large set of real images with fixed and known (\textit{n}) number of objects in each real image $\{\textbf{I}_s\}$ without any requirement of bounding box annotations. Since we know \textit{n} (in our case one object per image), while training $\mathcal{S}$ we can synthesize images with the same number of objects to pass to the discriminator, making it easier to train the generator. Having a single object image collection is not a requirement to train $\mathcal{S}$ and it has been shown in \cite{BlockGAN2020} that it can be trained successfully with 2 or more objects per image. 
However, having a large image collection $\{\textbf{I}_s\}$ with known number of objects is crucial for training $\mathcal{S}$. Our attempts to train it with a target image collection $\{\textbf{I}_t\}$ of city driving scenes, e.g. KITTI, with unknown number of objects per image were unsuccessful (details in supplementary material Sec. 4).

\vspace{-3mm}
\subsubsection{Obtaining Bounding Box Annotations}\label{sec:bbox_annot}
\vspace{-2mm}
The synthesis network $\mathcal{S}$ can generate a foreground object using a pose ($\boldsymbol{v}_f$, $\boldsymbol{l}_f$). This key property allows us to localize the object in the synthesized image and to create a 2D bounding box (BBox) annotation for it. We use the mean 3D bounding box (in real-world dimensions) of the object class and project it forward onto the 2D image plane 
using $\mathcal{S}$'s known camera matrix and the object's pre-defined pose ($\boldsymbol{v}_f$, $\boldsymbol{l}_f$) via perspective projection. 
The camera matrix is fixed for all synthesized images. We obtain the 2D bounding box $A_{g}$ for the synthesized image $I_{g}$ by computing the maximum and minimum coordinates of the projected 3D bounding box in the image plane. This procedure is illustrated in Fig.~\ref{fig:synthesis_net}. The paired data $\langle I_{g}, A_{g}\rangle$ can then be used to train the object detection network $\mathcal{F}$. 

\vspace{-1mm}
\subsection{Object Detection Adaptation}\label{sec:task_adapt}
\vspace{-1mm}
We introduce a set of objectives, which supervise $\mathcal{S}$ to synthesize images that are optimal for learning object detectors. These include an (a) object detection loss and (b) a multi-scale object synthesis loss, which we describe next.

\vspace{-4mm}
\subsubsection{Object Detection Loss}
\vspace{-2mm}
In our setup, we tightly couple the object detection network $\mathcal{F}$ to $\mathcal{S}$ such that it provides feedback to $\mathcal{S}$ (Fig.~\ref{fig:approach_overview}(b)). The object detection network $\mathcal{F}$ is a standard Feature Pyramid Network~\cite{Lin_2017_FPN}, which takes 2D images as input and predicts bounding boxes for the object. It is trained using the standard object detection losses ($\mathcal{L}_\mathit{det}$)~\cite{Lin_2017_FPN}. While training SSOD, we compute the object detection loss $\mathcal{L}_\mathit{det}$ for the image-annotation pairs $\langle\textbf{I}_{g}, \textbf{A}_{g}\rangle$ synthesized by $\mathcal{S}$ and use it as an additional loss term for updating the weights of $\mathcal{S}$. 

\vspace{-3mm}
\subsubsection{Multi-scale Object Synthesis Loss}
\vspace{-2mm}
It is important for $\mathcal{S}$ to be able to synthesize high quality images at varied object depths/scales, such that $\mathcal{F}$ can be optimally trained with diverse data. Hence, to extend the range of depths for which $\mathcal{S}$ generates high-quality objects, 
we introduce a multi-scale object synthesis loss, $\mathcal{L}_\mathit{mso}$ (Fig.~\ref{fig:approach_overview}(b)). 
To compute it, we use a synthesized image $I_{g}$'s bounding box $A_{g}$ and crop (in a differentiable manner) an image $I_{c}$ using a dilated version of $A_{g}$ with a unit aspect ratio such that the context around the object is considered. Further, we resize $I_{c}$ to $256 \times 256$. We then pass $I_{c}$ to a multi-scale object discriminator $\mathcal{D}_{mso}$. This makes the generated images match the appearance of the real images, with less surrounding background and simultaneously improves image quality. The real images input to $\mathcal{D}_{mso}$ are images from the source collection $\{\textbf{I}_{s}\}$, also of size $256 \times 256$. The multi-scale object synthesis loss, $\mathcal{L}_\mathit{mso}$ is then given by:
\vspace{-1mm}
\begin{equation}
	\mathcal{L}_\mathit{mso} = -\mathbb{E}_{I_c \sim p_\mathrm{synth}}[\mathcal{D}_{mso}(I_c)],
	\label{eqn:scn_loss}\vspace{-1mm}
\end{equation}

\noindent where $\mathcal{D}_{mso}(I_c)$ is the realism score predicted by $\mathcal{D}_{mso}$ for the image crop $I_c$.

\subsection{Target Data Adaptation} \label{sec:target_adapt}
\vspace{-1mm}
We train $\mathcal{S}$ with single-object images $\{\textbf{I}_{s}\}$ acquired from a collection, which do not necessarily come from the final target domain. Hence, there may be a domain gap between the images synthesized by $\mathcal{S}$ and those from the target domain (see examples in Fig.~\ref{fig:teaser} and Fig.~\ref{fig:qual_analysis}). This makes $\mathcal{F}$, trained on images synthesized by $\mathcal{S}$, perform sub-optimally on the target domain.
To address this, we introduce a target data adaptation module (Fig.~\ref{fig:approach_overview}(c)), whose focus is to adapt $\mathcal{S}$ such that it can synthesize images closer to the target data distribution. It uses foreground and background appearance losses to supervise training of $\mathcal{S}$, which make the synthesized images match the target domain. Additionally, it contains an object scale adaption block to match the scale of synthesised objects to the ones in the target domain. 
We align the synthesized data to the distribution of the target dataset without using any bounding box annotations. We describe these various components in detail.

\vspace{-4mm}
\subsubsection{Foreground Appearance Loss} 
\vspace{-2mm}
 We compute the foreground appearance loss $\mathcal{L}_\mathit{fg}$ via a patch-based~\cite{pix2pix2017} discriminator $\mathcal{D}_{fg}$ (Fig.~\ref{fig:approach_overview}(c)). It takes the synthesized image-annotation pair $\langle I_{g}, A_{g}\rangle$ as input and predicts a 2D class probability map, $\hat{c}_{fg}$ = $\mathcal{D}_{fg}(I_g)$, where $\hat{c}_{fg}$ is the patch-wise realism score for the synthesized image $I_g$.
 The foreground appearance loss ($\mathcal{L}_\mathit{fg}$) for the synthesis network $\mathcal{S}$ is given by:
\vspace{-1mm}
\begin{equation}
	\mathcal{L}_\mathit{fg} = -\mathbb{E}_{I_g \sim p_\mathrm{synth}}[{\hat{c}_{fg}}] * M_{g},
	\label{eqn:scn_loss}\vspace{-1mm}
\end{equation}
where $*$ indicates element-wise multiplication. $M_{g}$ masks the loss to be computed only for the foreground region of the synthesized image. 
The real images used to train this discriminator come from the target collection $\{\textbf{I}_{t}\}$. We acquire them by using the pre-trained object detection network $\mathcal{F}$ created during the first phase of uncoupled training (described in Sec.~\ref{sec:overview}). Specifically, we infer bounding boxes for the images in the target dataset $\{\textbf{I}_{t}\}$ using the pre-trained $\mathcal{F}$ and select a subset of images $\{\textbf{P}_{t}\}$ with detection confidence $>$0.9.
This forms an image-annotation pair $\langle P_{t}, M_{t}\rangle$, where $M_{t}$ is the corresponding binary mask for the detected foreground objects in image $P_{t}$. The loss for training the discriminator $\mathcal{D}_{fg}$ is computed as:
\vspace{-1mm}
\begin{equation}
	\mathcal{L}_\mathit{d_{fg}} = -\mathbb{E}_{I_t \sim p_\mathrm{real}}[c_t]*M_t + \mathbb{E}_{I_g \sim p_\mathrm{synth}}[{\hat{c}_{fg}}] * M_{g},
	\label{eqn:discr_loss}
	\vspace{-1mm}
\end{equation}
where $c_t$ is the patch-wise classification score predicted by $\mathcal{D}_{fg}$ for a real image.

\vspace{-3mm}
\subsubsection{Background Appearance Loss}
\label{sec:backgound_loss}
\vspace{-2mm}
The background discriminator $\mathcal{D}_{bg}$ is also a patch-based discriminator (Fig.~\ref{fig:approach_overview}(c)), which predicts the realism of the background region in $I_{g}$ with respect to the target data $\{\textbf{I}_{t}\}$. We compute the background mask by inverting the binary foreground mask $M_g$. The background appearance loss for the synthesis network, $\mathcal{S}$ is given by  
\vspace{-2mm}
\begin{equation}
	\mathcal{L}_\mathit{bg} = -\mathbb{E}_{I_g \sim p_\mathrm{synth}}[{\hat{c}_{bg}}] * (1 - M_{g}),
	\label{eqn:scn_loss}\vspace{-1mm}
\end{equation}
where $\hat{c}_{bg}$ = $\mathcal{D}_{bg}(I_g)$ predicts the patch-wise realism score for the background region of the generated image.

The real images used to train $\mathcal{D}_{bg}$ are obtained by identifying patches in the target collection $\{\textbf{I}_{t}\}$ where no foreground objects are present. To this end, we leverage pre-trained image classification networks and class-specific gradient-based localization maps using Grad-CAM~\cite{Selvaraju_gradcam}. Through this, we identify patches $\{\textbf{I}_{t}^{b}\}$ in the target collection $\{\textbf{I}_{t}\}$ that do not contain the object of interest. They serve as real samples of background images used to train $\mathcal{D}_{bg}$. The loss for training $\mathcal{D}_{bg}$ is computed as:
\begin{equation}
	\mathcal{L}_\mathit{d_{bg}} = -\mathbb{E}_{I_{t}^{b} \sim p_\mathrm{real}}[c_{t}^{b}] + \mathbb{E}_{I_g \sim p_\mathrm{synth}}[{\hat{c}_{bg}}] * (1 - M_{g}),
	\label{eqn:discr_loss}
	\vspace{-1mm}
\end{equation}
where $c_{t}^{b}$ is the patch-wise classification score predicted by $\mathcal{D}_{bg}$ for a real image. 

With $\mathcal{L}_{fg}$ and $\mathcal{L}_{bg}$ we only update the components of $\mathcal{S}$ that affect its style and appearance. These include (a) the parameters of the MLP blocks (Fig.~\ref{fig:synthesis_net}), which modify the foreground and background style codes and (b) the weights of 2D convolution layers. The foreground and background patches are obtained from the synthesized images using the annotations computed by our method (Sec. \ref{sec:bbox_annot}). Empirically, we observe
this is effective enough in learning the foreground and background distributions of the target domain.

\vspace{-2mm}
\subsubsection{Object Scale Adaptation}\label{sec:sinkhorn}
\vspace{-1mm}
We also find the optimal set of the object depth parameters that should be input into $\mathcal{S}$ to achieve the best performance on the target domain via this module.
To this end, we use  $\mathcal{S}$ to synthesize image-annotation pairs $\langle{I}_{g}^{d_r}, {A}_{g}^{d_r}\rangle$ for multiple different object depth ranges $\Theta$=$\{d_r\}$ and also obtain $\{\alpha^{d_r}\}$, which is the collection of cropped synthesized objects. Depth $d$ is one of the components of the location parameter $l$ used to specify the synthesized object's pose. We sample depth values uniformly within each depth range $d_r$. 
For each depth range $d_r$, we train a detector $\mathcal{F}^{d_r}$ with its corresponding synthetic data $\langle{I}_{g}^{d_r}, {A}_{g}^{d_r}\rangle$. We use $\mathcal{F}^{d_r}$ to detect all object bounding boxes $\{\beta^{d_r}\}$ in the target collection $\{\textbf{I}_{t}\}$, which have confidence $>$0.85. 
Finally we compute the optimal input depth interval for synthesis as:
\vspace{-1mm}
\small
\begin{equation}
	{d_o} = \operatorname*{argmin}_{d_i}  \mathcal{H}(\Phi(\alpha^{d_i}), \Phi(\beta^{d_i})),
	\label{eqn:sink_horn}\vspace{-1mm}%
\end{equation}
\normalsize
where $\Phi$ computes the conv5 features of a pre-trained image classification VGG~\cite{Simonyan15_vgg}) network and $\mathcal{H}$ is the Sinkhorn distance~\cite{NIPS2013_sinkhorn} between the two feature distributions. We use a single corresponding detector trained with the optimum depth range ${d_o}$ for the final evaluation on the target test data. 	

\subsection{Training Procedure}\label{sec:training_scheme}
\vspace{-1mm}
We adopt a stage-wise training strategy to learn SSOD. \\ 
\textbf{Uncoupled Training}.
We first pre-train $\mathcal{S}$ and $\mathcal{F}$ separately. We train the generator $\mathcal{S}$, supervised by the discriminators ${D}_{scn}$ and ${D}_{mso}$, using the source collection $\{\textbf{I}_{s}\}$ only. We then synthesize images with $\mathcal{S}$ containing 1 or 2 objects and compute their labels. We use them, along with real background regions extracted from the target data $\{\textbf{I}^{b}_{t}\}$ using Grad-CAM~\cite{Selvaraju_gradcam} (described in Sec.~\ref{sec:backgound_loss}) to pre-train $\mathcal{F}$. 

\vspace{1mm}
\noindent\textbf{Coupled Training}.
During this stage we couple all the networks together in an end-to-end manner and fine-tune them together with source $\{\textbf{I}_{s}\}$ and target $\{\textbf{I}_{t}\}$ collections, and the data synthesized by $\mathcal{S}$. We also adapt SSOD to the target data in this stage. We use a GAN-like training strategy and alternatively train $\mathcal{S}$ in one iteration and all other networks $\mathcal{D}_{scn}$, $\mathcal{F}$, $\mathcal{D}_{mso}$, $\mathcal{D}_{fg}$ and $\mathcal{D}_{bg}$ in the next one. Here $\mathcal{S}$ is supervised by all other modules and the total loss for training it is:
\vspace{-3mm}
\small

\begin{equation}
\begin{split}
  	\mathcal{L}_\mathit{syn} = & \lambda_{scn}\,\mathcal{L}_\mathit{scn} + \lambda_{mso}\,\mathcal{L}_\mathit{mso}+ \lambda_{det}\,\mathcal{L}_\mathit{det} \\
& +\lambda_{fg}\,\mathcal{L}_\mathit{fg} + \lambda_{bg}\,\mathcal{L}_\mathit{bg},
	\label{eqn:syn_loss1}  
 \end{split}
\end{equation}
\normalsize
\noindent where $\{\lambda_i\}$ are the relative weights of the various losses. Lastly, as discussed in Sec.~\ref{sec:sinkhorn} we find the optimal set of input object depth parameters for $\mathcal{S}$ that align synthesized data further to the target distribution.
\vspace{-1mm}
\section{Experiments}
\vspace{-2mm}
We validate SSOD for detecting ``car" objects in outdoor driving scenes. 
We assess quantitative performance using the standard mean Average Precision (mAP) metric at an Intersection-Over-Union (IOU) of 0.5.
We provide network architecture and training details in the supplementary.

\vspace{-1mm}
\subsection{Datasets and Evaluation}
\vspace{-2mm}
We use three datasets containing images of car objects to train and evaluate SSOD: (a) the Compcars dataset~\cite{yang2015large} as the single-car source dataset and (b) two multi-car KITTI~\cite{Geiger2012CVPR_kitti} and Cityscapes~\cite{Cordts_2016_CVPR_cityscapes} 
target datasets containing outdoor driving scenes. During training, we do not use bounding box annotations for any of these datasets.

\noindent\textbf{Compcars.}
The Compcars dataset~\cite{yang2015large} is an in-the-wild collection of 137,000 images with one car per image. 
It provides good diversity in car appearances, orientations and moderate diversity in scales (see examples Fig.~\ref{fig:teaser}). We use it as the source image collection $\{\textbf{I}_s\}$ to train our controllable viewpoint-aware synthesis network $\mathcal{S}$. 

\noindent\textbf{KITTI.} The challenging KITTI~\cite{Geiger2012CVPR_kitti} dataset contains $375\times1242$ sized outdoor driving scenes with zero or multiple cars per image with heavy occlusions, reflections and extreme lighting (see examples in Fig.~\ref{fig:teaser}). We use it as one of our target datasets $\{\textbf{I}_t\}$. We split it into disjoint \textit{training} (6000 unlabeled images) and \textit{validation} (1000 labeled images) sets. We report the mAP for Easy, Medium and Hard and all cases~\cite{Geiger2012CVPR_kitti} of the its \textit{validation} set.

\noindent\textbf{Cityscapes.} Similarly to KITTI, we also evaluate SSOD on the challenging Cityscapes~\cite{Cordts_2016_CVPR_cityscapes} outdoor driving target dataset with images of size $512\times1024$. We use the version provided by~\cite{cityscapes3d_nils} containing bounding box annotations. We split it into disjoint \textit{training} (3000 unlabeled images) and \textit{validation} (1000 labeled images) sets as provided in~\cite{cityscapes3d_nils}.

\begin{table*}\centering
	\begin{tabular}{lcccccccc}
		\toprule
		\multicolumn{1}{l}{Method} & Coupled & \multicolumn{1}{c}{Easy $\uparrow$} & \multicolumn{1}{c}{Medium $\uparrow$} & \multicolumn{1}{c}{Hard $\uparrow$} &  \multicolumn{1}{c}{All $\uparrow$} & \multicolumn{1}{c}{Sinkhorn~\cite{NIPS2013_sinkhorn} $\downarrow$} & \multicolumn{1}{c}{KID~\cite{demystifying_kid} $\downarrow$} & \multicolumn{1}{c}{FID~\cite{NIPS2017_8a1d6947_fid} $\downarrow$}\\
		\midrule
		\midrule
		BlockGAN~\cite{BlockGAN2020} 64                                                   & \textcolor{red}{\ding{55}}   & 65.1 & 48.3 & 40.5 & 51.3  & 0.486 & 0.048 & 8.3\\ 
		BlockGAN~\cite{BlockGAN2020} 128                                                  & \textcolor{red}{\ding{55}}     & 69.4  & 49.9 & 44.2  & 54.5 & 0.483 & 0.046 & 7.8\\
		BlockGAN~\cite{BlockGAN2020} 256                                                  & \textcolor{red}{\ding{55}}    & 72.7	 & 52.1 & 44.8  & 56.5 & 0.481 & 0.045 & 7.61\\
		\midrule
		SSOD \textit{w/o} {$\mathcal{L}_\mathit{fg}$ + $\mathcal{L}_\mathit{bg}$}        & \textcolor{green}{\checkmark}    & 74.7    & 59.3   & 52.7    & 62.2 & 0.475 & 0.042 &  7.22\\
		SSOD \textit{w/o} $\mathcal{L}_\mathit{mso}$                                      & \textcolor{green}{\checkmark}   & 78.3  & 65.6 & 53.5  & 65.8 & 0.471 & 0.040 & 6.86\\
		SSOD \textit{w/o} $OSA$                                                           & \textcolor{green}{\checkmark}    & 76.1  & 61.3 & 50.9  & 62.7 & 0.475 & 0.042 & 7.23\\
		\midrule
		SSOD-Full                                                                         & \textcolor{green}{\checkmark}     & \textbf{80.8}  & \textbf{68.1}  & \textbf{56.6}  & \textbf{68.4} & \textbf{0.465}  & \textbf{0.037} & \textbf{6.37}
		   
		\\
		\midrule
	\end{tabular}
	\caption{\textbf{Ablation study on KITTI.} Rows 1-3: BlockGAN in $\mathcal{S}$ trained without coupling to the detector at different image resolutions; rows 4-6: different ablated versions of SSOD each with one component removed; and row 7: full SSOD model. Columns 1-3: mAP value at IOU 0.5 for KITTI's Easy, Medium, Hard and All cases; and columns 4-6: Sinkhorn, KID, and FID scores to compare object regions in synthesized and real-world KITTI images.
	}

	\label{tab::kitti_ablation}
	\vspace{-4mm}
\end{table*}

\vspace{-1mm}
\subsection{Ablation Study}\label{sec:ablation_study}
\vspace{-2mm}
We conduct ablation studies on the KITTI dataset to evaluate the contribution of each individual component of SSOD (Table~\ref{tab::kitti_ablation}). We evaluate object detection performance using mAP, and compute SinkHorn~\cite{NIPS2013_sinkhorn}, KID~\cite{demystifying_kid} and FID~\cite{NIPS2017_8a1d6947_fid} scores to compare the appearance of the synthesized foreground objects to objects in KITTI. 

\noindent\textbf{Quality of annotations} Firstly, we estimate the accuracy of annotations obtained from our pipeline. For 260 images synthesized by the generator, we manually annotate the bounding boxes and measure the mAP between them and
the annotations by our pipeline. It is 0.95 at an IoU of 0.5, which is
reasonable for learning object detectors.

\begin{figure}
	\centering
	\includegraphics[width=\columnwidth]{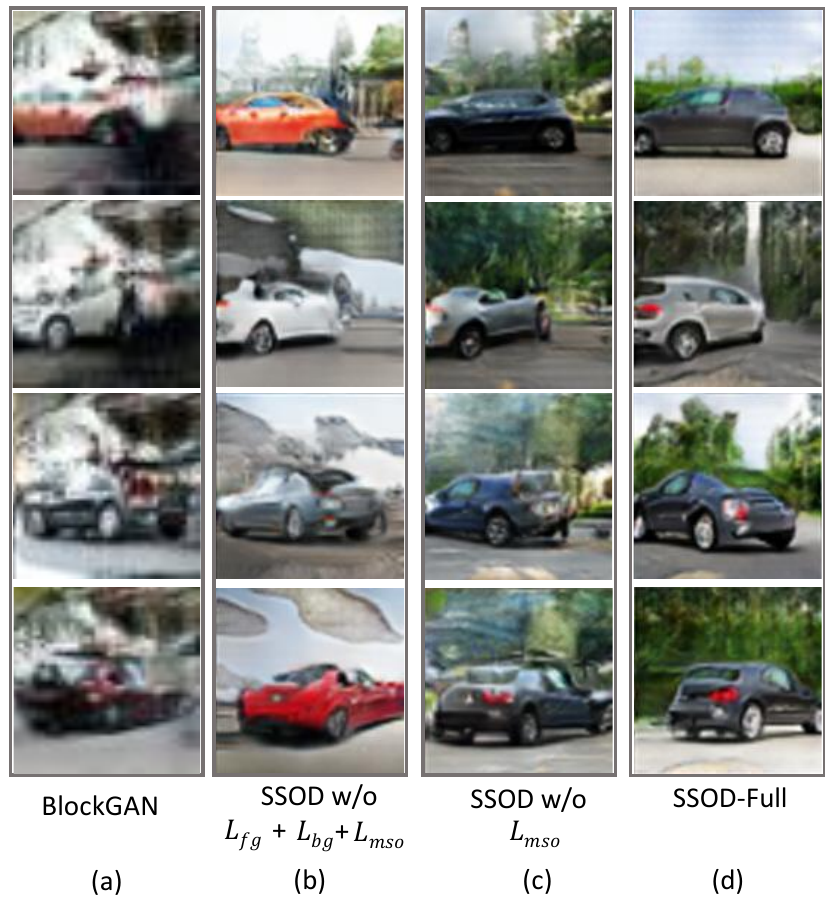}
	\caption{\textbf{Qualitative analysis of image synthesis.} The columns show images generated by (a) BlockGAN~\cite{BlockGAN2020} at $64\times 64$; and by $\mathcal{S}$ for (b) SSOD trained without $\mathcal{L}_\mathit{fg}$,  $\mathcal{L}_\mathit{bg}$, and $\mathcal{L}_\mathit{mso}$; (c) SSOD trained without $\mathcal{L}_\mathit{mso}$; and (d) the full SSOD model. Each row has images generated with the same pose, and foreground and background style codes. Rows (b)-(d) show $256\times 256$ sized images.
}
	\label{fig:qual_analysis}
	\vspace{-5mm}
\end{figure}

\vspace{1mm}
\noindent\textbf{Uncoupled Training.} We evaluate the efficacy of simply training the object detector $\mathcal{F}$ with images synthesized by $\mathcal{S}$, when each of these networks is trained separately without coupling.
We compare the original BlockGAN~\cite{BlockGAN2020} with an image resolution of $64\times64$ to two of its variants with image resolutions $128\times128$ and $256\times256$ that we train as described in Sec.~\ref{sec:synth_net}.
The results are shown in the top three rows of Table~\ref{tab::kitti_ablation}. They indicate that synthesized foreground objects at higher resolutions improve the Sinkhorn, KID and FID metrics, which, in turn, translate to corresponding gains in the object detector's performance as well. The improvements in visual quality achieved by higher resolution synthesis are also evident in the first two columns of Fig.~\ref{fig:qual_analysis}. We further observed that training the detector without background target images found with Grad-CAM results in false positive detections and reduces mAP from 56.5 to 51.6. 

\vspace{1mm}
\noindent\textbf{Coupled Training.} Next, we evaluate the performance of variants of SSOD trained with  coupled synthesis ($\mathcal{S}$) and object detection ($\mathcal{F}$) networks. We evaluate four variants of SSOD: (a) without the target data appearance adaption losses described in Sec.~\ref{sec:target_adapt} (SSOD \textit{w/o} $\mathcal{L}_\mathit{fg}$ + $\mathcal{L}_\mathit{bg}$); (b) without the multi-scale object synthesis loss $\mathcal{L}_\mathit{mso}$ described in Sec.~\ref{sec:task_adapt} (SSOD \textit{w/o} $\mathcal{L}_\mathit{mso}$); (c) without adaptation to the target dataset's object scales as described in Sec.~\ref{sec:target_adapt} (SSOD \textit{w/o} $OSA$); and (d) the full SSOD model (SSOD-full).
We observe that, across the board, all variants of SSOD trained with a coupled  detector (bottom four rows of Table~\ref{tab::kitti_ablation}) perform significantly better than those without (top three rows of Table~\ref{tab::kitti_ablation}). This result verifies the usefulness of our proposed end-to-end framework, which adapts the synthesis network $\mathcal{S}$ to both the downstream task of object detection as well as to the target dataset's distribution. The best performance, overall, is achieved by our full SSOD model with the highest mAP score of $68.4$. Removing each of our individual proposed modules for target data appearance adaptation (SSOD \textit{w/o} $\mathcal{L}_\mathit{fg}$ + $\mathcal{L}_\mathit{bg}$), target object scale adaptation (SSOD \textit{w/o} $OSA$) and multi-object scale synthesis (SSOD \textit{w/o} $\mathcal{L}_\mathit{mso}$) from SSOD-Full result in a reduction in its performance, with the target data appearance adaption model affecting SSOD's detection accuracy the most.

\noindent\textbf{Qualitative Analysis.} We qualitatively evaluate the effect of our proposed losses on the images synthesized by $\mathcal{S}$. In each row of  Fig.~\ref{fig:qual_analysis} we show images synthesized with the same foreground and background style codes, but with variants of the network $\mathcal{S}$ trained with a different set of losses in each column. Columns 2-4 are at a resolution of $256 \times 256$. We vary the foreground and background style codes across the rows. All objects are synthesized at a large depth from the camera.
Fig.~\ref{fig:qual_analysis}(a) shows the images synthesized by the original BlockGAN~\cite{BlockGAN2020} at a resolution of $64\times64$ suffers from poor quality.
Fig.~\ref{fig:qual_analysis}(b) shows the synthesized images by our method when trained with the coupled object detector at higher resolution, leads to better visibility. By adding target data appearance adaptation losses ($\mathcal{L}_\mathit{fg}$ + $\mathcal{L}_\mathit{bg}$), images (Fig.~\ref{fig:qual_analysis}(c)) match the appearance of target distribution. Finally, adding the multi-scale object synthesis loss $\mathcal{L}_\mathit{mso}$ leads to the best result (high visual quality and appearance alignment to the target distribution). These qualitative results corroborate with their quantitative counterparts: Sinkhorn, KID and FID metrics in Table~\ref{tab::sota_comparison}. 

\vspace{-2mm}
\subsection{Comparisons to State-of-the-Art}
\vspace{-2mm}
On the KITTI dataset, we compare SSOD to existing methods, Wetectron~\cite{ren-eccv2020_wetectron2} and PCL~\cite{tang2018pcl},  capable of training object detectors without requiring bounding box annotations. These methods similar to 
SSOD, train object detectors solely with unlabeled image collections. They also do not use 3D CAD models and hence are the most directly comparable methods to SSOD. Wetectron~\cite{ren-eccv2020_wetectron2} is the best-performing prior method. We train Wetectron and PCL with a combination of Compcars~\cite{yang2015large} and KITTI's~\cite{Geiger2012CVPR_kitti} \textit{training} set; use image-level labels for the presence/absence of the object; get object proposals from Edgeboxes~\cite{zitnick2014edgeboxes}; and evaluate it on KITTI's \textit{validation} set. The results are in Table~\ref{tab::sota_comparison}. Compared to Wetectron (mAP of 38.1 for All) and PCL (mAP of 33.2 for All), SSOD (mAP of 68.4 for All) has $\sim$2X better detection accuracy. We believe that SSOD's superior performance results from its use of a pose-aware synthesizer to generate data for training object detectors.
The GAN improves the training data's diversity and also optimally adapts to the task of object detection on target data.

\setlength{\tabcolsep}{1.5pt} 
\small
\begin{table}\centering
	\begin{tabular}{llccccc}
		\toprule
		& \multicolumn{1}{l}{Method} & \multicolumn{1}{l}{3D Assets} & \multicolumn{1}{c}{Easy$\uparrow$} & \multicolumn{1}{c}{Medium$\uparrow$} & \multicolumn{1}{c}{Hard$\uparrow$} & \multicolumn{1}{c}{All$\uparrow$} \\
		\midrule
		\small
		\scriptsize
		& PCL~\cite{tang2018pcl}                   		 & \textcolor{green}{\ding{55}} & 47.3    & 32.9   & 19.4    & 33.2\\

		& Wetectron~\cite{ren-eccv2020_wetectron2}                    		 & \textcolor{green}{\ding{55}} & 51.3    & 37.9   & 25.1    & 38.1\\

		& SSOD-Full (ours)                    			 & \textcolor{green}{\ding{55}} & \textbf{80.8}  & \textbf{68.1}  & \textbf{56.6}  & \textbf{68.4} \\
		\midrule
		& Meta-Sim\color{Black}{\footnotemark}~\cite{kar2019metasim}                       			 & \textcolor{red}{\checkmark} & 65.9  & 66.3  & 66.0  & 66.0 \\
		& Meta-Sim2~\cite{devaranjan2020metasim2}      & \textcolor{red}{\checkmark} & 67.0  & 67.0  & 66.2  & 66.7 \\

		\midrule
	\end{tabular}
	\caption{\textbf{Comparisons to SOTA.} Object detection performance (mAP at IOU 0.5) on KITTI of SSOD and various SOTA methods.} 
	\label{tab::sota_comparison}
	\vspace{-6mm}
\end{table}
\footnotetext{We report detection accuracy values for the version of Meta-Sim that does not use labeled validation images from the KITTI~\cite{Geiger2012CVPR_kitti} dataset.}

\normalsize

We also compare SSOD to SOTA rendering-based methods Meta-Sim~\cite{kar2019metasim} and Meta-Sim2~\cite{devaranjan2020metasim2}. They train object detectors purely using synthetically rendered data and evaluate on unlabeled real-world datasets. They require large libraries of 3D CAD models and hence use strong geometric priors. 
In contrast, SSOD does not use any 3D CAD assets. In fact, its synthesis network can be viewed as a controllable renderer learned only from object image collections without geometric priors. Interestingly, even without using any strong geometric priors, SSOD surpasses both Meta-Sim and Meta-Sim2 for Easy, Medium and All cases in KITTI (Table~\ref{tab::sota_comparison}). For Hard cases, SSOD performs lower than Meta-Sim and Meta-Sim2, mostly due its low image quality for occluded objects and its lower 2D bounding box label precision (see Sec.~\ref{sec:failure_cases}).
Nevertheless, it is exciting that even without using 3D assets and by merely learning from image collections, SSOD can compete with rendering-based methods, which require significant supervision. 

\vspace{-1mm}
\subsection{Additional Dataset}
\vspace{-2mm}
An advantage of SSOD is that it can adapt to different target datasets. To validate this, we additionally evaluate it's performance on Cityscapes~\cite{Cordts_2016_CVPR_cityscapes}. We evaluate the full SSOD model trained on Compcars and Cityscapes; its ablated versions with specific individual components removed (as described in Sec.~\ref{sec:ablation_study} -- Coupled Training); BlockGAN in $\mathcal{S}$ not coupled with the detector and trained with Compcars only; and the competing Wetectron method trained on Compcars and Cityscapes (Table~\ref{tab::citscapes}). Similar to KITTI, for Cityscapes too, SSOD-Full achieves the best performance (mAP of 31.3). Removing {$\mathcal{L}_\mathit{fg}$ + $\mathcal{L}_\mathit{bg}$}, which help adapt SSOD to Cityscapes, affects its performance the most. All variants of SSOD jointly trained with the detector perform better than the uncoupled BlockGAN in $\mathcal{S}$. SSOD-Full also performs significantly better than Wetectron (mAP of 18.2).

\setlength{\tabcolsep}{5pt} 
\small
\begin{table}\centering
	\begin{tabular}{llcc}
		\toprule
		& \multicolumn{1}{l}{Method} & \multicolumn{1}{c}{mAP$\uparrow$} & \multicolumn{1}{c}{Sinkhorn$\downarrow$}\\
		\midrule
		&Wetectron~\cite{ren-eccv2020_wetectron2}         & 18.2 & 0.549\\
		&BlockGAN~\cite{BlockGAN2020} 256                       & 22.7 & 	0.531\\
		\hline
		& SSOD \textit{w/o} {$\mathcal{L}_\mathit{fg}$ + $\mathcal{L}_\mathit{bg}$} 	& 27.2  &  		0.520 \\
		& SSOD \textit{w/o} $\mathcal{L}_\mathit{mso}$                      				& 28.5 	&  	0.515 \\
		& SSOD \textit{w/o} $OSA$                   						& 29.1  &   0.514 \\
		\midrule
		& SSOD-Full                         	   						& \textbf{31.3}   &  \textbf{0.506}\\
		\midrule
	\end{tabular}
	\caption{\textbf{Performance on Cityscapes.} Object detection performance (mAP at IOU 0.5) and synthetic data quality analysis (Sinkorn) on Cityscapes.} 
	\label{tab::citscapes}
	\vspace{-2mm}
\end{table}

\normalsize

\vspace{-1mm}
\subsection{Discussion on Results}
\label{sec:failure_cases}
\vspace{-2mm}
SSOD suffers from low recall for the Hard cases in KITTI as it fails to detect heavily occluded cars (examples in supplementary material). Fig.~\ref{fig:plot_pr} shows SSOD's precision-recall curves on KITTI for IOU thresholds: 0.5 (solid) and 0.45 (dashed). Also, with a lower IOU threshold of 0.45 its mAP improves: 80.8 to 83.5 (Easy), 68.1 to 73.2 (Medium) and 56.6 and 63.6 (Hard). This indicates that improving the precision of the synthesized objects' bounding boxes labels can lead to improvements in SSOD's performance.

\begin{figure}
	\centering
	\includegraphics[width=0.5\columnwidth]{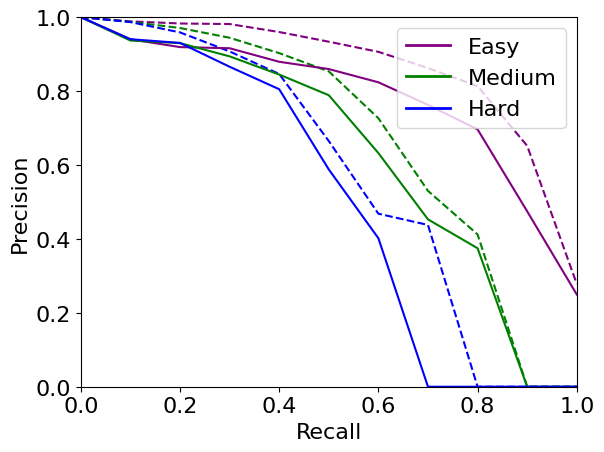}
	\caption{\textbf{Precision-recall curves on KITTI.} Curves for SSOD with IOU thresholds of 0.5 (bold lines) and 0.45 (dashed lines).	}
	\label{fig:plot_pr}
	\vspace{-4mm}
\end{figure}
\vspace{-2mm}
\section{Conclusion}
\vspace{-2mm}
SSOD is the first work to leverage controllable GANs to learn object detectors in a self-supervised manner with unlabelled image collections. It not only opens up an exciting new research paradigm in the area, but also shows that significant detection accuracy can be achieved by using controllable image synthesis. 
Controllable GANs are able to synthesize data with diversity and realism to train object detectors. They also allow the flexibility to adapt them optimally via end-to-end training to the downstream detection task and target domains. 
With the rapid progression of controllable GANs, we envision that the gains acquired there would lead to further improvements on GAN-based self-supervised object detection. 

{\small
\bibliographystyle{ieee_fullname}
\bibliography{egbib}
}

\clearpage

\setcounter{section}{0}
\renewcommand{\thesection}{\Alph{section}}
\section*{\Large Appendix}

\section{Overview}
In this supplementary document, we provide the architectural and training details of our SSOD framework. Section \ref{sec:architecture} provides the architectures for the various networks ($\mathcal{S}$, $\mathcal{F}$, $\mathcal{D}_{scn}$, $\mathcal{D}_{fg}$, $\mathcal{D}_{bg}$) used in SSOD. Section \ref{sec:training} describes the training procedure and all its hyper-parameters. Further, in Section \ref{sec:ablations} we present the analysis of the case when the pose-aware synthesis network is trained directly on the target data with unknown number of objects per image, which fails to successfully disentangle the background and foreground representations. Lastly, we discuss the performance of the object detector on heavily occluded objects (`Hard' cases in KITTI~\cite{Geiger2012CVPR_kitti}) in Section \ref{sec:discussion}. We open-source our code at \url{https://github.com/NVlabs/SSOD}.

\section{Network Architectures}\label{sec:architecture}

\subsection{Pose-Aware Synthesis Network}
The architecture for the pose-aware synthesis network ($\mathcal{S}$), described in Section 3.3 and Figure 3 of the main paper, is detailed in Table~\ref{tab::pas_net}. The architectures of $\mathcal{S}$'s foreground and background object branches are presented in Table~\ref{tab::fg_branch} and Table~\ref{tab::bg_branch}. As shown in Figure 3 of the main paper, each of these branches take a learnable 3D object representation as input and perform a series of 3D convolutions on them. The 3D features are stylized using separate input style codes for the foreground ($z_f$) and the background ($z_b$) objects. The style codes pass through their respective MLP blocks, the output of which is used by adaptive instance normalization (AdaIN)~\cite{huang2017adain} to control the style of each generated object/background. The MLP blocks contain 4 fully-connected layers each with a width of 200 neurons and leaky ReLU activations after each layer. The resulting 3D features are finally transformed using the input pose. These branches are used by $\mathcal{S}$ (Table~\ref{tab::pas_net}), which performs an element-wise maximum operation on all the 3D features from all foreground objects and the background, followed by a projection of these features onto 2D. This is further followed by a set of 2D convolutions to generate an image of size $256\times256$.

\subsection{Scene and MSO Discriminator}
The architectures of scene ($\mathcal{D}_{scn}$) and multi-scale object ($\mathcal{D}_{mso}$) discriminators are described in Table~\ref{tab::discr_net}. The input to each of these networks is an image of size $256\times256$. This is followed by a set of 2D convolutions and a fully-connected layer to produce a single class membership score indicating the probability of real/fake for the input image. We use Spectral Normalization~\cite{miyato2018spectral} in these discriminator networks.

\subsection{Patch Discriminator for Foreground and Background}
The patch-based architectures of the foreground ($\mathcal{D}_{fg}$) and the background ($\mathcal{D}_{bg}$) discriminators are described in Table~\ref{tab::patch_net}. The input to each of these networks is an image of size $256\times256$. Each network is fully convolutional and produces an output of size $8\times8$. We use Spectral Normalization~\cite{miyato2018spectral} in these discriminator networks.

\subsection{Object Detection Network}
We use Faster-RCNN~\cite{renNIPS15fasterrcnn} with a Resnet-50-FPN~\cite{Lin_2017_FPN} backbone as our detection network. It takes a 2D image as input and extracts features using the backbone layer. These features are further used by the object detection head to predict the top-left and bottom-right corners of the bounding box pertaining to a detected object. We use the object detection implementation from~\cite{wu2016tensorpack} in our work.

\setlength{\tabcolsep}{8pt} 
\begin{table*}\centering
	\begin{tabular}{rcccccc}
		\toprule
		& Layer & Kernel Size & stride & Activation & Normalization & Output Dimension \\ 
		\toprule
		\midrule
		& Learnable 3D-Code  	  &  - 	   &	-	 	& 	 LReLU	  		&		  AdaIN    	  & $4\times4\times4\times512$   \\ 
		\midrule
		& 3D-Deconv &  $3 \times 3 \times 3$ & 2 & LreLU & AdaIN & $8\times8\times8\times128$   \\ 
		\midrule
		& 3D-Deconv &  $3 \times 3 \times 3$ & 2 & LreLU & AdaIN & $16\times16\times16\times64$ \\ 
		\midrule	
		& 3D-Transform &  - & - & - & - & $16\times16\times16\times64$ \\ 
		\midrule		
	\end{tabular}
	\caption{\textbf{Architecture of foreground object branch}}
	\label{tab::fg_branch}
	\vspace{-2mm}
\end{table*}

\setlength{\tabcolsep}{8pt} 
\begin{table*}\centering
	\begin{tabular}{rcccccc}
		\toprule
		& Layer & Kernel Size & stride & Activation & Normalization & Output Dimension \\ 
		\toprule
		\midrule
		& Learnable 3D-Code  	  &  - 	   &	-	 	& 	 LReLU	  		&		  AdaIN    	  & $4\times4\times4\times256$   \\ 
		\midrule
		& 3D-Deconv &  $3 \times 3 \times 3$ & 2 & LreLU & AdaIN & $8\times8\times8\times128$   \\ 
		\midrule
		& 3D-Deconv &  $3 \times 3 \times 3$ & 2 & LreLU & AdaIN & $16\times16\times16\times64$ \\ 
		\midrule		
		& 3D-Transform &  - & - & - & - & $16\times16\times16\times64$ \\ 
		\midrule		
	\end{tabular}
	\caption{\textbf{Architecture of background object branch}}
	\label{tab::bg_branch}
	\vspace{-2mm}
\end{table*}

\setlength{\tabcolsep}{8pt} 
\begin{table*}\centering
	\begin{tabular}{rcccccc}
		\toprule
		& Layer & Kernel Size & stride & Activation & Normalization & Output Dimension \\ 
		\toprule
		\midrule
		\parbox[t]{4mm}{\multirow{3}{*}{\rotatebox[origin=c]{90}{3D Branches}}}

		& $n \times$ FG Branch (a)  &  - 	   &	-	 	& 	 -	  		&		  -    	  & $16\times16\times16\times64$   \\ 
    	\tiny		&  			  	  &     &  	 		&		    & 	  		&	   	\\ 
		\cdashline{2-7}
    	\tiny		&  			  	  &     &  	 		&		    & 	  		&	   	\\ 
		& BG Branch (b)  	  		&  - 	   &	-	 	& 	 -	  		&		  -    	  & $16\times16\times16\times64$   \\ 
    	\tiny		&  			  	  &     &  	 		&		    & 	  		&	   	\\ 
		\midrule
		& Element-Wise Maximum(a,b) 	&  - 	   & 	- 		& 	 - 			& 		  - 	  & $16\times16\times16\times64$   \\ 
		\midrule
		\rule{0pt}{3ex}		
		\parbox[t]{4mm}{\multirow{2}{*}{\rotatebox[origin=c]{90}{Project}}}		
		& Collapse 	&  - 	   & 	- 		& 	 - 			& 		  - 	  & $16\times16\times(16.64)$   \\ 
		& Conv 	&  $1 \times 1$ 	   & 	1 		& 	 LReLU 			& 		  - 	  & $16\times16\times256$   \\ 
		\small
		&  			  	  &     &  	 		&		    & 	  		&	   	\\ 
		\midrule
		\parbox[t]{4mm}{\multirow{6}{*}{\rotatebox[origin=c]{90}{2D Convs}}}		
		& 2D-Deconv 	&  $4 \times 4$ 	   & 	2 		& 	 LReLU 			& 		  - 	  & $32\times32\times128$   \\ 
		\cline{2-7}		
		& 2D-Deconv 	&  $4 \times 4$ 	   & 	2 		& 	 LReLU 			& 		  - 	  & $64\times64\times64$   \\ 
		\cline{2-7}		
		& 2D-Deconv 	&  $4 \times 4$ 	   & 	2 		& 	 LReLU 			& 		  - 	  & $128\times128\times32$   \\ 
		\cline{2-7}		
		& 2D-Deconv 	&  $4 \times 4$ 	   & 	2 		& 	 LReLU 			& 		  - 	  & $256\times256\times16$   \\ 
		\cline{2-7}						
		& 2D-Conv (to RGB)		&  $4 \times 4$ 	   & 	1 		& 	 -	 			& 		  - 	  & $256\times256\times3$   \\ 
		\midrule						
	\end{tabular}
	\caption{\textbf{Architecture of pose-aware synthesis network}}
	\label{tab::pas_net}
	\vspace{-2mm}
\end{table*}

\setlength{\tabcolsep}{8pt} 
\begin{table*}\centering
	\begin{tabular}{rcccccc}
		\toprule
		& Layer & Kernel Size & stride & Activation & Normalization & Output Dimension \\ 
		\toprule
		\midrule	
		& 2D-Conv (from RGB)	&  $4 \times 4$ 	   & 	1 		& 	 LReLU 			& 	-	   	  & $256\times256\times8$   \\ 
		\midrule		
		& 2D-Conv 	&  $5 \times 5$ 	   & 	2 		& 	 LReLU 			& 		  SpectralNorm 	  & $128\times128\times16$   \\ 
				\midrule
		& 2D-Conv 	&  $5 \times 5$ 	   & 	2 		& 	 LReLU 			& 		  SpectralNorm 	  & $64\times64\times32$   \\ 
		\midrule
		& 2D-Conv 	&  $5 \times 5$ 	   & 	2 		& 	 LReLU 			& 		  SpectralNorm 	  & $32\times32\times64$   \\ 
		\midrule
		& 2D-Conv 	&  $5 \times 5$ 	   & 	2 		& 	 LReLU 			& 		  SpectralNorm 	  & $16\times16\times128$   \\ 
		\midrule		
		& 2D-Conv 	&  $5 \times 5$ 	   & 	2 		& 	 LReLU 			& 		  SpectralNorm 	  & $8\times8\times256$   \\ 
		\midrule		
		& 2D-Conv 	&  $5 \times 5$ 	   & 	2 		& 	 LReLU 			& 		  SpectralNorm 	  & $4\times4\times512$   \\ 
		\midrule		
		& Fully Connected 	&  - 	   & 	- 		& 	 - 			& 		-   	  & 1   \\ 
		\midrule								
	\end{tabular}
	\caption{\textbf{Architecture of scene discriminator and multi-scale object discriminators}}
	\label{tab::discr_net}
	\vspace{-2mm}
\end{table*}

\setlength{\tabcolsep}{8pt} 
\begin{table*}\centering
	\begin{tabular}{rcccccc}
		\toprule
		& Layer & Kernel Size & stride & Activation & Normalization & Output Dimension \\ 
		\toprule
		\midrule	
		& 2D-Conv (from RGB) 	&  $4 \times 4$ 	   & 	1 		& 	 LReLU 			& 		   -	  & $256\times256\times8$   \\ 
		\midrule		
		& 2D-Conv 	&  $4 \times 4$ 	   & 	2 		& 	 LReLU 			& 		  SpectralNorm 	  & $128\times128\times16$   \\ 
		\midrule
		& 2D-Conv 	&  $4 \times 4$ 	   & 	2 		& 	 LReLU 			& 		  SpectralNorm 	  & $64\times64\times32$   \\ 
		\midrule
		& 2D-Conv 	&  $4 \times 4$ 	   & 	2 		& 	 LReLU 			& 		  SpectralNorm 	  & $32\times32\times64$   \\ 
		\midrule
		& 2D-Conv 	&  $4 \times 4$ 	   & 	2 		& 	 LReLU 			& 		  SpectralNorm 	  & $16\times16\times128$   \\ 
		\midrule		
		& 2D-Conv 	&  $4 \times 4$ 	   & 	2 		& 	 LReLU 			& 		  SpectralNorm 	  & $8\times8\times256$   \\ 
		\midrule		
		& 2D-Conv 	&  $1 \times 1$ 	   & 	1 		& 	 LReLU 			& 		  - 	  & $8\times8\times256$   \\ 
		\midrule		
	\end{tabular}
	\caption{\textbf{Architecture of foreground and background discriminators}}
	\label{tab::patch_net}
\end{table*}

\section{Training}\label{sec:training}
\vspace{-3mm}
We implement SSOD in Tensorflow~\cite{tensorflow}. We use the Adam~\cite{kingma2015adam}  optimizer to learn our networks with beta parameter values (0.9, 0.99). The learning rate for all the networks is set to 0.00005. The batch size is 16. The dimensions of the style codes for the foreground ($z_f$) and the background ($z_b$) are 200 and 100, respectively. The style codes are sampled from a uniform distribution between (-1, 1). For our camera in $\mathcal{S}$ we use a focal length of $35mm$  with a sensor size of $32mm$ similar to~\cite{BlockGAN2020}.

\subsection{Training Procedure}
As described in the paper, we adopt a stage-wise training strategy to learn the modules of SSOD. 

\noindent\textbf{Uncoupled Training.} Here, we first train the pose-aware synthesis network for 20 epochs. We initialize the weights of $\mathcal{S}$, $\mathcal{D}_{scn}$, $\mathcal{D}_{mso}$ using $\mathcal{N}(0, 0.2)$ and biases with $0$. We train $\mathcal{S}$, supervised by the discriminators $\mathcal{D}_{scn}$ and $\mathcal{D}_{mso}$ in a Generative Adversarial Network~\cite{NIPS2014_gans} framework. We found empirically that updating $\mathcal{S}$ twice for every update of $\mathcal{D}_{scn}$ and $\mathcal{D}_{mso}$, results in the best visual quality. The weights for the losses from $\mathcal{D}_{scn}$ and $\mathcal{D}_{mso}$ are 0.5, each. The real images for $\mathcal{D}_{scn}$ and $\mathcal{D}_{mso}$ are sampled from the real-world source image collection $\{\textbf{I}_{s}\}$. During this training stage, we synthesize images with a single foreground object as described in Sec. 3.3 of the main paper. 

Next, we train the object detector $\mathcal{F}$ (initialized with ImageNet pretraining) using 10k images synthesized by $\mathcal{S}$ containing 1 or 2 objects paired with their computed bounding box labels $\langle \textbf{I}_{g}, \textbf{A}_{g}\rangle$.  The learning rate for $\mathcal{F}$ is set to 0.00005 and it is trained for 10 epochs. In addition to the images synthesized from $\mathcal{S}$, we use real background regions $\{\textbf{I}_{t}^{b}\}$ extracted from target collection $\{\textbf{I}_{t}\}$. To obtain the real background images $\{\textbf{I}_{t}^{b}\}$, we leverage Grad-CAM~\cite{Selvaraju_gradcam} to identify regions in $\{\textbf{I}_{t}\}$ that do not contain the object of interest (`Vehicle', `Car', `Wagon', `Van') with a high confidence ($>0.9$). To achieve this, we use layer-4 of the Resnet-152 network trained on Imagenet with Grad-CAM.

\noindent\textbf{Coupled Training.} In this stage, we tightly couple all networks ($\mathcal{S}$, $\mathcal{F}$, $\mathcal{D}_{scn}$, $\mathcal{D}_{mso}$, $\mathcal{D}_{fg}$, $\mathcal{D}_{bg}$) together in an end-to-end manner and fine-tune them with the source $\{\textbf{I}_{s}\}$, target $\{\textbf{I}_{t}\}$ and synthesized $\{\textbf{I}_{g}\}$ images. Similar to the uncoupled training stage, the real images for training $\mathcal{D}_{scn}$ and $\mathcal{D}_{mso}$ are sampled from the source collection $\{\textbf{I}_{s}\}$. The real images for training $\mathcal{D}_{bg}$ come from $\{\textbf{I}_{t}^{b}\}$.

The real images for training $\mathcal{D}_{fg}$ are obtained as follows. We use the object detector $\mathcal{F}$ trained in the uncoupled training stage to obtain high-confidence $(>0.9)$ detections of cars in $\{\textbf{I}_{t}\}$. Further, we use these detections to crop out image patches $\{\textbf{P}_{t}\}$ of size 256$\times$256 around the object using the centers of the detections. These form image-annotation pairs $\langle P_t, M_t \rangle$ where $M_t$ is the corresponding binary mask indicating the region inside the detection. $M_t$ is used in computing the foreground appearance loss as discussed in Sec. 3.5.1 of the main paper.

 The weights for the losses from $\mathcal{D}_{fg}$ and $\mathcal{D}_{bg}$, ($\mathcal{L}_{fg}$ and $\mathcal{L}_{bg}$) are initially set to a 0.05 and are progressively increased by 0.01 every 200 iterations to reach 0.5. $\mathcal{L}_{fg}$ and $\mathcal{L}_{bg}$ update the components of $\mathcal{S}$ that affect the overall appearance of images. Hence only the parameters of the MLPs and of 2D convolutional blocks of $\mathcal{S}$ are updated. $\mathcal{F}$ is trained using images synthesized by $\mathcal{S}$ using the image-annotation pairs $\langle I_g, A_g \rangle$. The weight for $\mathcal{L}_{det}$ is 0.4. We train SSOD in coupled training stage for 25k iterations.

 \section{Training synthesizer directly on target data}\label{sec:ablations}
\begin{figure*}[t]
	\centering
	\includegraphics[width=0.8\textwidth]{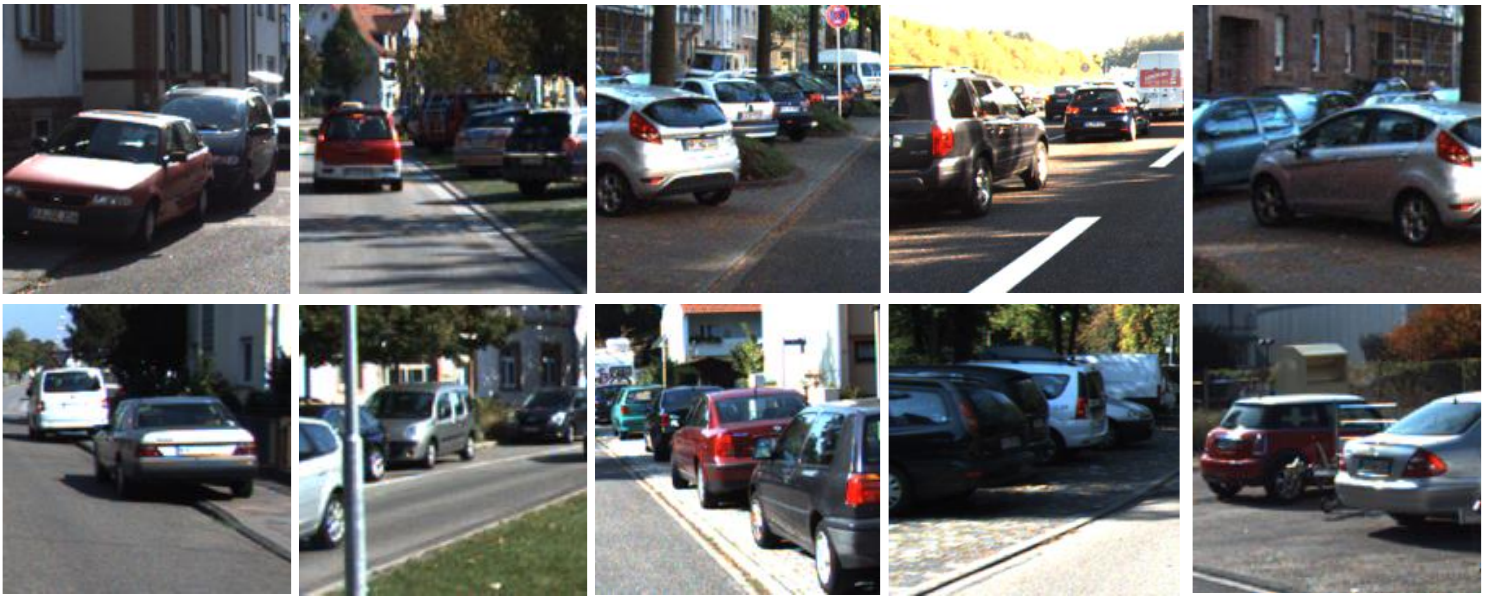}
	\caption{\textbf{Sample image from target data (KITTI).} The figure shows sample images from KITTI~\cite{Geiger2012CVPR_kitti} data obtained for training the synthesis network $\mathcal{S}$. They are obtained by using Grad-CAM~\cite{Selvaraju_gradcam} on the original KITTI images and finding regions where there is a high confidence response for the classes `Vehicle', `Car', `Wagon', `Van'. It can be seen that each image contains a random number of multiple objects and not necessarily a single object.}
	\label{fig:kitti_data}
	\vspace{-4mm}
\end{figure*}

\vspace{-2mm}
In this section, we explore the case when the pose-aware synthesis network, $\mathcal{S}$ is trained directly on the target data $\{\textbf{I}_t\}$ instead of the source data $\{\textbf{I}_s\}$. As described in Sec. 4.1 of the main paper, we use the Compcars dataset~\cite{yang2015large} as the source dataset $\{\textbf{I}_s\}$, which is an in-the-wild collection of images with one car per image (see examples in Figure 1 of main paper). The target dataset for this experiment is the KITTI~\cite{Geiger2012CVPR_kitti} dataset which contains outdoor driving scenes with unkown numbers of cars image (zero to multiple cars per image) with heavy occlusions, reflections and extreme lighting (see examples in Figure~\ref{fig:kitti_data}). 

\vspace{1mm}
To train $\mathcal{S}$ with the target data, we first identify regions in $\{\textbf{I}_t\}$, which contain foreground objects of interest. We use Grad-CAM~\cite{Selvaraju_gradcam} to identify regions in the target image collections where there is a high confidence response for the classes (`Vehicle', `Car', `Wagon', `Van'). We do this to obtain training images with a fixed number (one) car per image.  Figure~\ref{fig:kitti_data} illustrates some example images obtained using this method. However, notice that these images do not necessarily contain only one foreground object, but a random unknown number of them per image. This is because Grad-CAM can only identify regions, which have a high response to a specific class and cannot separate out objects. However, it is important for $\mathcal{S}$ to know the number of foreground objects per image as discussed in Sec. 3.3 of the main paper. 

\begin{figure*}[t]
	\centering
	\includegraphics[width=0.8\textwidth]{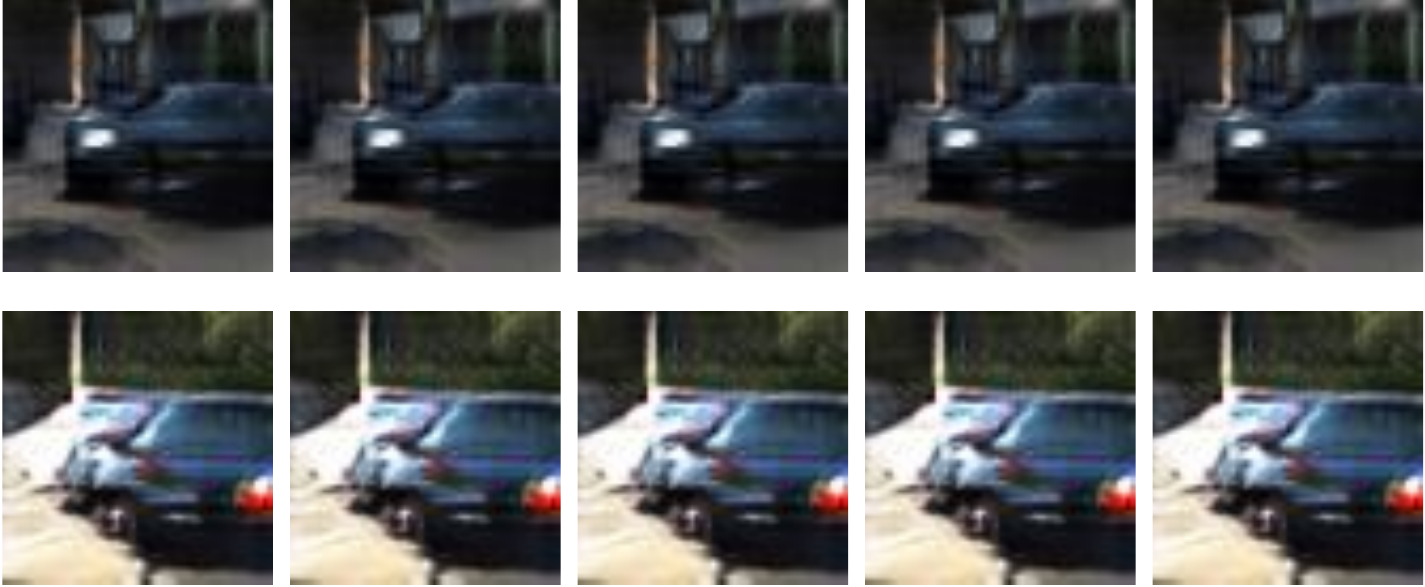}
	\caption{\textbf{Images synthesized from $\mathcal{S}$ trained on the target image collection from KITTI (Figure~\ref{fig:kitti_data}).} Each row in this figure contains images with a fixed input foreground and background code. The input horizontal translation of the foreground objects is varying across the columns. It can be seen that even when the input translation of the foreground object varies, the image is constant across the columns. This is because $\mathcal{S}$ is unable to disentangle the foreground and background regions and hence translating the foreground object does not induce any changes in the generated images.}
	\label{fig:kitti_results}
	\vspace{-4mm}
\end{figure*}

\vspace{1mm}
Since the number of foreground objects is not known \textit{a priori} for these real-world images, it is not possible to correctly specify the number of foreground objects in the synthesizer while synthesizing images with it. This makes it difficult for $\mathcal{S}$ to disentangle foreground and background regions and learn separable representations for them during training with such data. Figure~\ref{fig:kitti_results} presents images synthesized by $\mathcal{S}$ trained on foreground images extracted from the KITTI target image collection (Figure~\ref{fig:kitti_data}). Each row in this figure contains images with a fixed input foreground and background code. The horizontal translation value of the foreground objects is varying across the columns. Figure~\ref{fig:kitti_results} shows that, even when the input translation of the foreground object varies, the image is constant throughout the columns without translation of the foreground object. This is because $\mathcal{S}$ is unable to disentangle the foreground and background regions and hence providing a different input translation value to the GAN for the foreground object does not induce any changes in the generated images. 

\begin{figure*}[t!]
	\centering
	\includegraphics[width=0.8\textwidth]{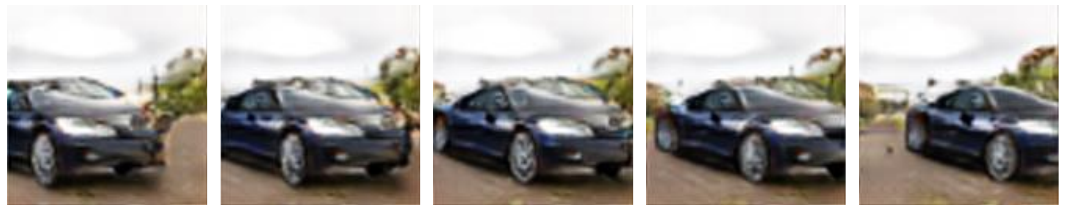}
	\caption{\textbf{Images synthesized from $\mathcal{S}$ trained on source image collection.} The input horizontal translation of the foreground object is varying from left to right.
	The object is translating along the horizontal axis according to the input translation, while the background remains constant. This is because on being trained with the source image collection $\mathcal{S}$ is able to disentangle the foreground and background representations.}
	\label{fig:car_tx}
	\vspace{-4mm}
\end{figure*}

On the other hand, when $\mathcal{S}$ is trained using a source dataset $\{\textbf{I}_s\}$ (where the number of objects per image is known) and adapted to a target dataset (KITTI~\cite{Geiger2012CVPR_kitti}) using our approach, it is able to disentangle foreground and background representations. As a result, the pose of the foreground object can be controlled by the input parameters as shown in Figure.~\ref{fig:car_tx}.
This experiment shows that it is imperative to have access to an image collection $\{\textbf{I}_s\}$ where the number of foreground objects per image is known a priori to be able to successfully train our controllable GAN network $\mathcal{S}$. In our case, we assume that we have access to an image collection with one object per image.

\vspace{-3mm}
\section{Limitations}\label{sec:discussion} 
\begin{figure*}[th]
	\centering
	\begin{subfigure}{\textwidth}
		\centering
		\includegraphics[width=0.95\textwidth]{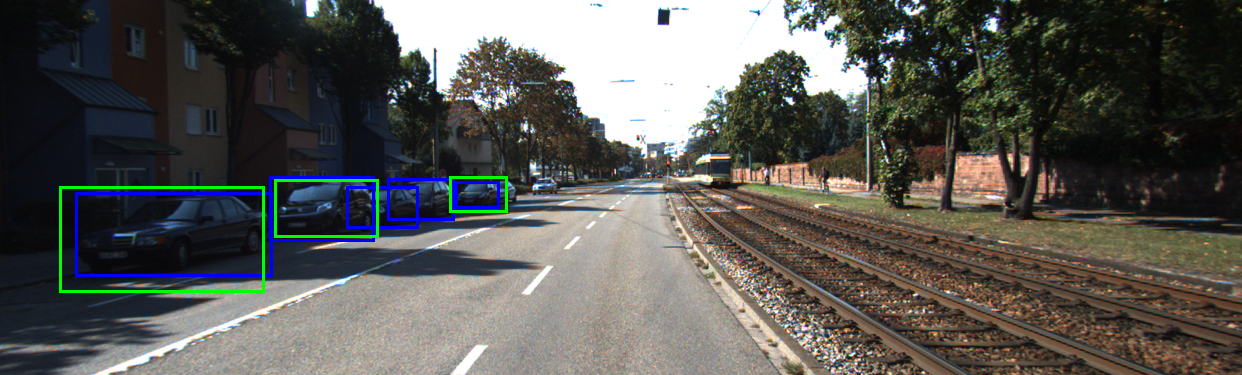}
	\end{subfigure}
	\par\smallskip
	
	\begin{subfigure}{\textwidth}
		\centering
		\includegraphics[width=0.95\textwidth]{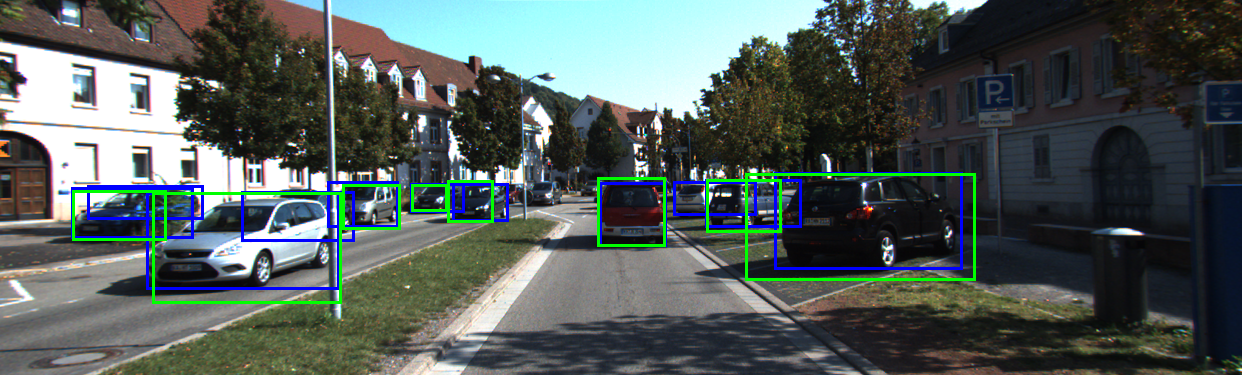}
	\end{subfigure}
	\par\smallskip
	
	\begin{subfigure}{\textwidth}
		\centering
		\includegraphics[width=0.95\textwidth]{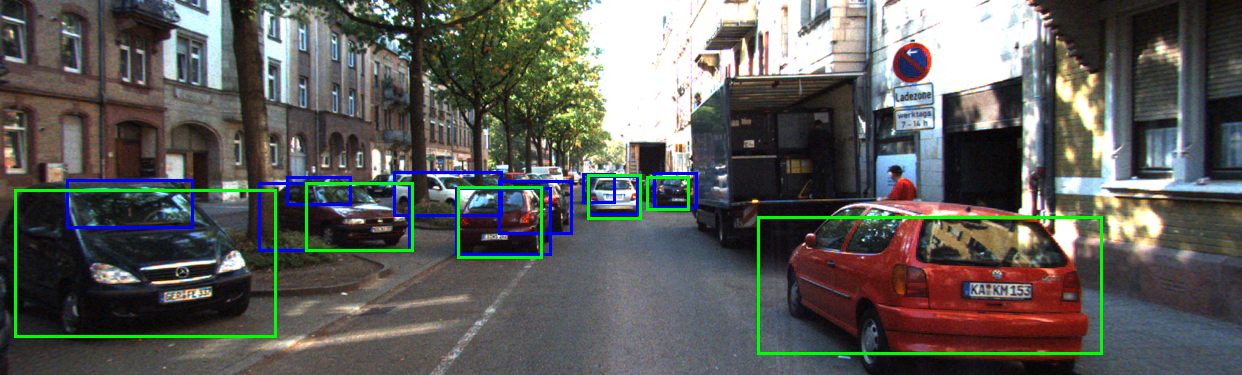}
	\end{subfigure}
	\par\smallskip
	
	\begin{subfigure}{\textwidth}
		\centering
		\includegraphics[width=0.95\textwidth]{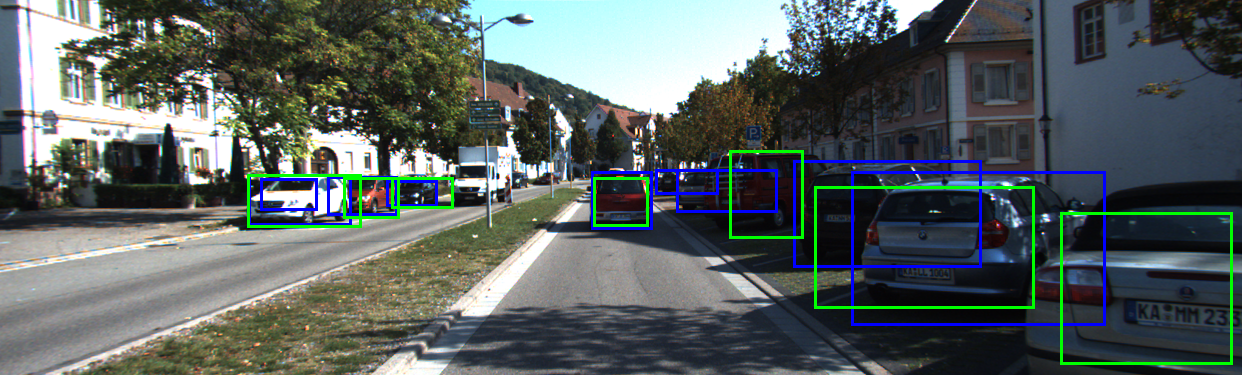}
	\end{subfigure}
	
	\caption{\textbf{Visualization of object detections by SSOD on the KITTI~\cite{Geiger2012CVPR_kitti} dataset.} \textcolor{green}{Green} boxes show SSOD's predictions and \textcolor{blue}{Blue} boxes show ground truth annotations. It can be seen that most of the cars with none to moderate occlusions are reliably detected. Cars with extremely heavy occlusions where only a small part of the car can be seen are sometimes missed.}
	\label{fig:vis1}
\end{figure*}

\begin{figure*}[th]
	\centering
	\begin{subfigure}{\textwidth}
		\centering
		\includegraphics[width=0.95\textwidth]{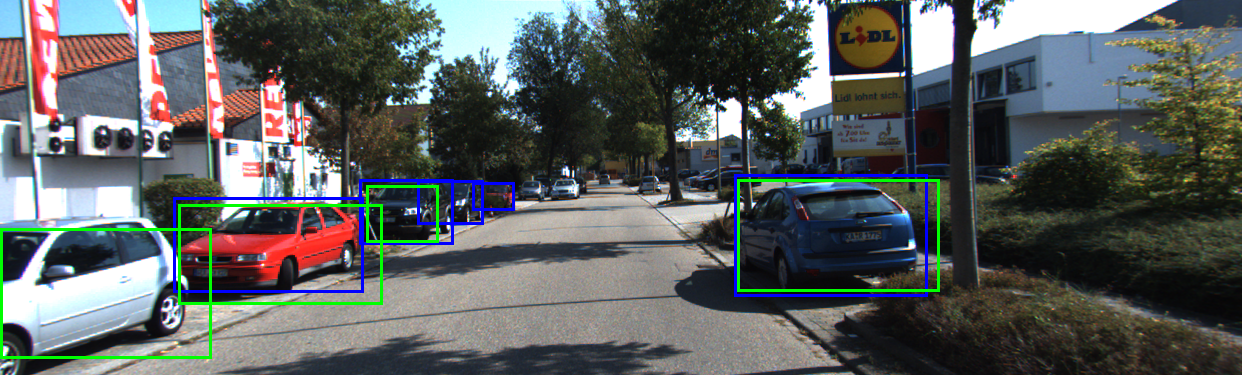}
	\end{subfigure}
	\par\smallskip
	
	\begin{subfigure}{\textwidth}
		\centering
		\includegraphics[width=0.95\textwidth]{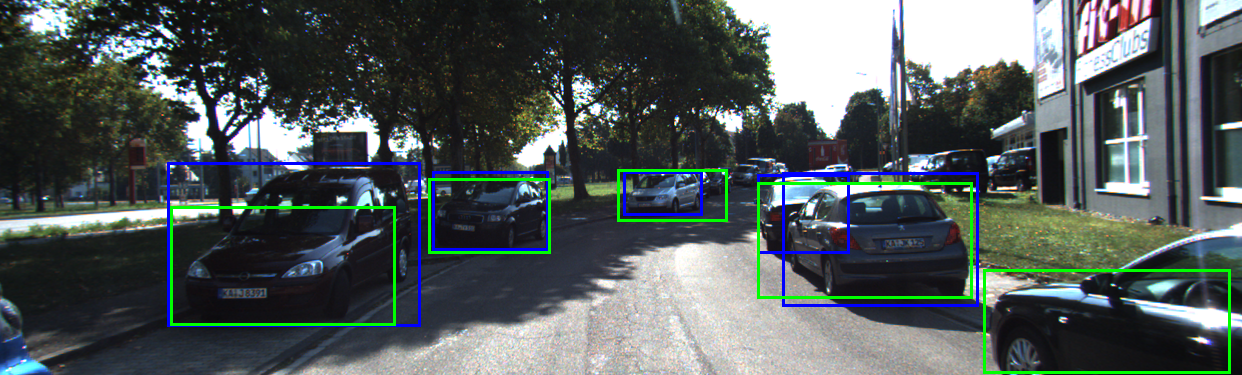}
	\end{subfigure}
	\par\smallskip
	
	\begin{subfigure}{\textwidth}
		\centering
		\includegraphics[width=0.95\textwidth]{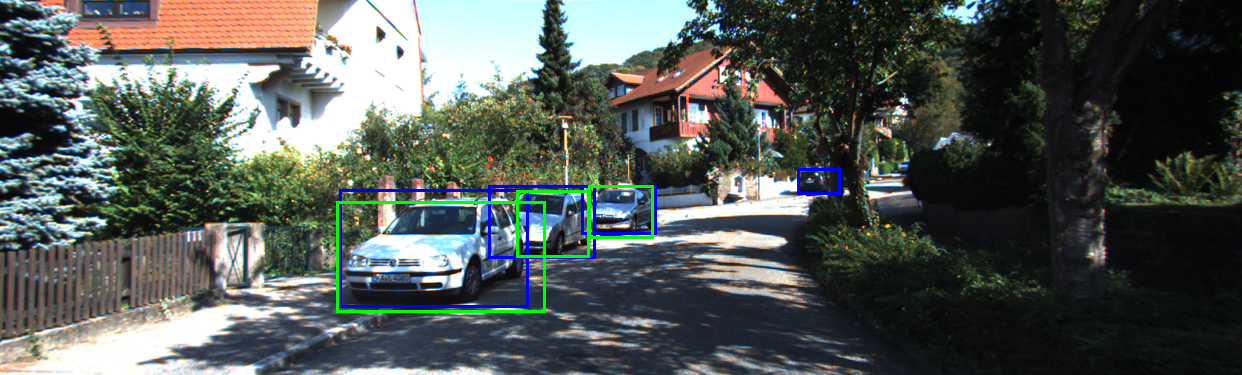}
	\end{subfigure}
	\par\smallskip
	
	\begin{subfigure}{\textwidth}
		\centering
		\includegraphics[width=0.95\textwidth]{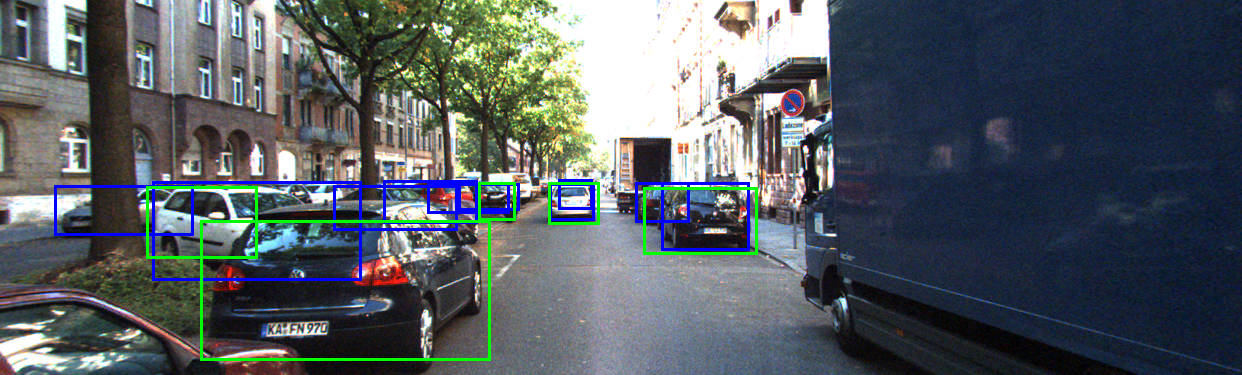}
	\end{subfigure}
	
	\caption{\textbf{Visualization of object detections by SSOD on the KITTI~\cite{Geiger2012CVPR_kitti} dataset.} \textcolor{green}{Green} boxes show SSOD's predictions and \textcolor{blue}{blue} boxes show ground truth annotations. It can be seen that most of the cars with none to moderate occlusions are reliably detected. Cars with extremely heavy occlusions where only a small part of the car can be seen are sometimes missed.}
	\label{fig:vis2}
\end{figure*}

\vspace{-1mm}
In this section, we present a qualitative analysis of the objects detected by SSOD for images from the challenging KITTI~\cite{Geiger2012CVPR_kitti} dataset as discussed in Sec. 4.5 of the main paper. These images contain objects with heavy occlusions, reflections and extreme lighting variations. In Figures~\ref{fig:vis1}, \ref{fig:vis2} we illustrate the 2D bounding boxes predicted by SSOD in \textcolor{green}{green} and the ground truth bounding boxes in \textcolor{blue}{blue}. In all these images it can be seen that the cars are reliably detected under moderate occlusions, lighting variations and reflections. On the other hand, cars with extremely heavy occlusions 
where only a small part of the car is visible are sometimes missed. We observe that these are generally cases where cars are parked along the side of the road and are heavily occluded by each other. Such cases where there is heavy occlusion are classified as `Hard' in the KITTI~\cite{Geiger2012CVPR_kitti} dataset. We believe that this problem can be alleviated by specifically learning to model the layout, occlusions of objects and context of scenes in generated images to be similar to target data. For example, cars could occlude each other in parked scenarios or roads with heavy traffic. We consider this to be a future goal to address.

\clearpage


\end{document}